\documentclass[11pt]{article}

\usepackage{arabtex}
\usepackage{utf8}

\usepackage[final]{acl}
\usepackage{booktabs}

\usepackage{lipsum}
\usepackage{pdfpages}
\usepackage{enumitem}

\usepackage{caption}
\usepackage{times}
\usepackage{latexsym}
\usepackage[T1]{fontenc}
\usepackage{microtype}
\usepackage[noupquote]{inconsolata}

\usepackage[utf8]{inputenc}
\usepackage{todonotes}
\usepackage{multicol}

\usepackage[most]{tcolorbox}
\usepackage{amssymb}
\usepackage{pdflscape}
\usepackage{graphicx}

\title{Benchmarking Arabic Dialects for Math Problems}

\author{
\textbf{Noor Abo Mokh,\textsuperscript{1}
Teresa Lynn,\textsuperscript{1}
Nizar Habash,\textsuperscript{1,2}
Ahmed Elshabrawy,\textsuperscript{1}}\\
\textbf{Sara Alhajeri,\textsuperscript{1}
Alaa Elsetohy,\textsuperscript{1}
Preslav Nakov,\textsuperscript{1}
Alham Fikri Aji\textsuperscript{1}}\\
\textsuperscript{1}Mohamed bin Zayed University of Artificial Intelligence,\\
\textsuperscript{2}New York University Abu Dhabi\\
\texttt{\{noor.abomokh,alham.fikri\}@mbzuai.ac.ae} }

\usepackage{epstopdf} 

\makeatletter
\let\orig@latex@error\@latex@error
\long\def\@latex@error#1#2{%
  \edef\reserved@a{\detokenize{#1}}%
  \def\reserved@b{Loading a class or package in a group}%
  \edef\reserved@b{\detokenize\expandafter{\reserved@b}}%
  \ifx\reserved@a\reserved@b
    \typeout{Bypassed ArabTeX group check for: #1}%
  \else
    \orig@latex@error{#1}{#2}%
  \fi
}
\makeatother

\begin{document}

\maketitle
\begin{abstract}
Mathematical reasoning remains underexplored for low-resource languages, including Dialectal Arabic. To bridge this gap, we introduce the DAR-MGSM (Dialectal Arabic Multilingual Grade School Math) Benchmark, the first manually curated benchmark for evaluating mathematical reasoning in Dialectal Arabic. DAR-MGSM consists of math word problems (MWPs) requiring multi-step reasoning and covers five major Arabic dialects alongside Modern Standard Arabic (MSA). The benchmark comprises two parallel versions:  \textit{translated-only} and \textit{translated \& culturally localized}. The latter  version adapts entities and references to regional varieties.
We evaluate eleven Arabic-centric and multilingual large language models on DAR-MGSM. The results show that the performance generally declines on the \textit{localized} version, highlighting the challenges introduced by culturally grounded adaptations. However, the magnitude of this effect varies across models, with Arabic-centric models exhibiting different levels of sensitivity to localization. 
A mechanistic interpretability analysis shows that translated and localized variants of the same MWPs often induce distinct internal reasoning trajectories, even when aggregate accuracy remains similar.
Overall, our results show that mathematical reasoning performance depends on both linguistic variety and cultural context.
\end{abstract}

\setcode{utf8}
\section{Introduction}

Large language models (LLMs) have demonstrated strong performance on complex tasks like mathematical reasoning in resource-rich languages such as English \citep{cobbe2021gsm8k,hendrycks2021measuringmassivemultitasklanguage,hendrycks2021measuringmathematicalproblemsolving} and French \citep{hoang-ha-2025-pensez}. However, these capabilities do not consistently extend to low-resource languages \citep{sobhani2026mathmist,huang2025benchmax}, nor to languages with numerous variants.
Arabic is a prominent example: despite its rich dialectal diversity, most existing 
evaluation mathematical benchmarks focus 
on Modern Standard Arabic (MSA).

\begin{figure}
    \centering
    \includegraphics[width=0.95\linewidth]{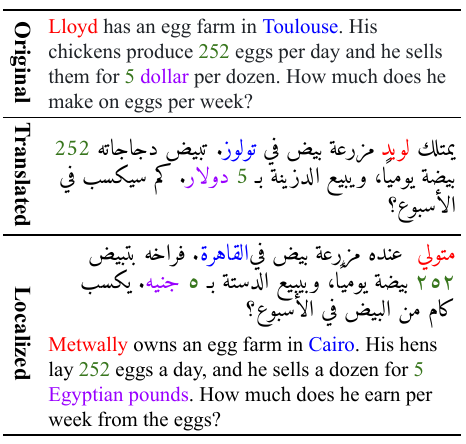}
    \caption{An illustrative Math Word Problem adapted from MGSM-Rev2 in three versions: (\emph{i})~Original English, (\emph{ii})~Translated to MSA, and (\emph{iii})~Localized to Egyptian Arabic dialect and entities.}
    \label{fig:intro-example}
    \vspace{-10pt}
\end{figure}

The focus on MSA is understandable. Math textbooks and educational materials across the Arab world are written in MSA. Yet, teachers use Dialectal Arabic (DA) extensively in instruction. \citet{amin2020teaching} document the extensive use of dialect in science classes, while MSA is used in textbooks. Other studies such as \citet{amara1996arabic} mention extensive use of dialect in math classes to explain concepts.

This reflects Arabic's diglossic nature \cite{Ferguson:1959:diglossia}, an critical  challenge for LLM evaluation. Existing benchmarks focusing on MSA provide only a partial view of mathematical reasoning capabilities in Arabic. To bridge this gap, we aim to develop an LLM math word problem (MWP) benchmark 
that reflects classroom real-world conditions in the Arab world.

Leveraging existing benchmarks, we translate MWPs from English into Arabic, 
to further allow us to investigate LLM biases across MSA, and Arabic dialects. While translation captures linguistic differences, it often creates texts alien to Arab culture. American English MWPs, for instance, commonly use US-centric entities such as Western personal names and locations, or currencies, even when presented to Arabic-speaking users.\footnote{For example, US dollars for a MWP set in New York vs. US dollars for a MWP set in Egypt.} See Figure~\ref{fig:intro-example} for an illustrative example.

Such choices influence how naturally problems are interpreted and processed. 

To investigate these challenges, we distinguish between two orthogonal benchmark dimensions: \emph{linguistic translation}, which adapts the language of a problem, and \emph{cultural entity alignment or localization}, which adapts culturally grounded entities for names, locations, currencies, and numeral conventions. This distinction enables a controlled study of whether culturally aligned benchmark variants elicit different model behavior than direct translations of the same mathematical problem.

Additionally, we benchmark localization with mechanistic interpretability, providing insight not only into \emph{whether} localization affects performance, but also \emph{how} it influences internal model reasoning processes.

Our contributions are as follows:
\begin{itemize}

\item We introduce DAR-MGSM (Dialectal Arabic Multilingual Grade School Math Benchmark), the first benchmark for evaluating grade-school mathematical reasoning in MSA and five major Arabic varieties: Egyptian (EGY), Emirati (UAE), Jordanian (JOR), Moroccan (MOR), and Saudi (KSA).

\item We release two parallel benchmark variants: a manually translated dataset and a translated \& culturally localized dataset.

\item We conduct the first systematic evaluation of dialectal Arabic mathematical reasoning, benchmarking multiple Arabic-centric and multilingual LLMs.

\item We use mechanistic interpretability to compare model behavior on translated and localized data.

\end{itemize}

Our results show that performance generally declines on localized data compared to direct translations, with varying effects across a variety of models and dialects. The interpretability analysis reveals that localization alters LLM reasoning, highlighting the need for culturally grounded evaluation in Arabic and other multilingual settings, where translation alone may mask differences in model capabilities.

\section{Related Work}

\paragraph{Mathematical reasoning benchmarks}
Evaluation of LLMs on mathematical reasoning has expanded substantially beyond English in recent years. One of the widely used benchmarks is GSM8K \citep{cobbe2021gsm8k} containing approximately 8,500 grade-school MWPs. Building on GSM8K, Shi et al.~\shortcite{shi2022languagemodelsmultilingualchainofthought} introduced MGSM, extending to ten additional languages translated from English. MGSM-Rev2 \cite{cobbe2021gsm8k,shi2022languagemodelsmultilingualchainofthought,peter2025mindgapnottranslation} is a revised version of MGSM. MGSM8KInstruct \citep{chen-etal-2024-breaking}, is a multilingual instruction-following dataset built on MGSM. Other reasoning benchmarks, such as GPQA \citep{rein2023gpqa}, remain predominantly English-centric.
In the above benchmarks, Arabic is typically represented only through MSA,
but not dialects. We bridge this gap by evaluating on language varieties, providing parallel grade-school math benchmarks for MSA and five major Arabic dialects.

\paragraph{Arabic mathematical reasoning}
Within the Arabic NLP landscape, mathematical reasoning resources remain comparatively limited. Multilingual evaluation suites such as BenchMAX \citep{huang2025benchmax} and MathMist \citep{sobhani2026mathmist} include MSA as one of several evaluation languages. More recently, \citet{alenezi2026mathar} introduced large-scale Arabic mathematical reasoning datasets, including AGSM8K, Qudurat, and ArabicMWPs, where AGSM8K consists of GSM8K-style problems translated into MSA. ArMATH \citep{alghamdi-liang-zhang:2022:LREC} provides a collection of manually authored MWPs in MSA designed to support the development of automatic MWP solvers. Other dialectal benchmarks focus on related reasoning tasks, such as multiple-choice question answering \citep{altakrori-etal-2026-dialectalarabicmmlu}, rather than open-ended mathematical problem solving. To the best of our knowledge, no benchmark currently evaluates grade-school mathematical reasoning across the Arabic dialects. Consequently, the effect of dialectal variation on LLM mathematical reasoning capabilities remains largely unexplored. In contrast, we 
provide a benchmark specifically designed for evaluation 
covering multiple dialects. 
\paragraph{Translation vs. localization}
A methodological limitation shared by many multilingual reasoning benchmarks is a reliance on direct translation from English. For example, MGSM was intentionally translated without localization to facilitate cross-lingual comparison \citep{shi2022languagemodelsmultilingualchainofthought}. Similarly, AGSM8K \citep{alenezi2026mathar} relies on machine translation followed by manual review.
Recent work has argued that translation alone is insufficient for culturally grounded evaluation. Studies in multilingual and culturally aware NLP demonstrate that direct translations can obscure important sociocultural differences and reduce linguistic naturalness, motivating the need for localization in benchmark design \citep{farsi-etal-2025-melac,zhou2025culture}. This issue is particularly relevant for Arabic, where dialects are closely tied to specific regional and cultural contexts.
Our work builds on these insights by explicitly separating \emph{translation} and \emph{localization} as two distinct dimensions of the benchmark. Unlike prior Arabic mathematical reasoning benchmarks, 
we provide both translated and culturally localized versions of the same MWPs. This design enables controlled evaluation of the impact of localization on mathematical reasoning performance while preserving the underlying problem structure.

\paragraph{Multilingual reasoning challenges}
Recent studies have shown that multilingual reasoning remains challenging for current LLMs. Models often achieve higher accuracy when reasoning in English than when reasoning directly in other languages, reflecting the predominance of English in training and optimization pipelines \citep{saji2026reasoning}. Particularly, \citet{saji2026reasoning} demonstrate that reasoning in English yields higher final-answer accuracy than reasoning in many other languages.
These findings motivate our investigation of localized Arabic MWPs, where linguistic variation and cultural alignment can be studied independently. By comparing translated and localized variants of the same MWP, we isolate the effect of cultural grounding on mathematical reasoning performance and better understand how benchmark design influences model behavior.

\paragraph{Residual stream interpretability}
Mechanistic interpretability often treats the transformer residual stream as a shared computational workspace, where attention heads and MLPs read from and write to common subspaces \citep{elhage2021mathematical}. Methods such as 
tuned lens probe intermediate residual representations, while causal tracing, activation patching, and path patching test whether specific hidden states or model components are necessary for a given behavior \citep{belrose2023tuned,meng2022locating,olsson2022context,nanda2023progress}. 
\begin{table*}[t]
\centering
\small
\begin{tabular}{lp{13cm}}
\toprule
\textbf{ENG} &
\texttt{\textless name\textgreater} has an egg farm in
\texttt{\textless location\textgreater}. His chickens produce
\texttt{\textless num1\textgreater} eggs per day and he sells them for
\texttt{\textless num2\textgreater} \texttt{\textless currency\textgreater}
per dozen. How much does he make on eggs per week? \\
\midrule
\textbf{MSA} &
\multicolumn{1}{r}{
\<يمتلك \LR{<name>} مزرعة بيض في \LR{<location>}. تبيض دجاجاته \LR{<num1>} بيضة يوميًا،>}\\
&
\multicolumn{1}{r}{\<ويبيع الدزينة بـ \LR{<currency>} \LR{<num2>}. كم سيكسب في الأسبوع؟> }\\ \midrule

\textbf{EGY} &
\multicolumn{1}{r}{
\<\LR{<name>} عنده مزرعة بيض في \LR{<location>}. فراخه بتبيض \LR{<num1>} بيضة يوميًا،>}\\
&
\multicolumn{1}{r}{
\<وبيبيع الدستة بـ \LR{<currency>} \LR{<num2>}. ييكسب كام من البيض في الأسبوع؟>}\\
\midrule

\textbf{JOR} &
\multicolumn{1}{r}{
\<\LR{<name>} عنده مزرعة بيض في \LR{<location>}. جاجاته ببيضوا \LR{<num1>} بيضة في اليوم،>}\\
&
\multicolumn{1}{r}{
\<وببيع الدزينة منهم ب \LR{<currency>} \LR{<num2>}. قديش بيطلع دخله من البيض بالأسبوع؟>}\\
\midrule

\textbf{KSA} &
\multicolumn{1}{r}{
\<\LR{<name>} عنده مزرعة بيض في \LR{<location>}. دجاجاته تبيض \LR{<num1>} بيضه في اليوم>}\\
&
\multicolumn{1}{r}{
\<ويبيعها ب \LR{<currency>} \LR{<num2>} لكل درزن. كم مكسبه في الأسبوع؟>}\\
\midrule

\textbf{MOR} &
\multicolumn{1}{r}{
\<\LR{<name>} عندو فيرما ديال الدجاج في \LR{<location>}. الدجاج ديالو كيبيض \LR{<num1>} بيضة ف النهار،>}\\
&
\multicolumn{1}{r}{
\<كيبيع البيض ب \LR{<currency>} \LR{<num2>} للزينة. شحال كيربح من البيض ف كل سيمانا؟>}\\
\midrule

\textbf{UAE} &
\multicolumn{1}{r}{
\<عند \LR{<name>} مزرعة دواجن في \LR{<location>}. ديايه يبيضون \LR{<num1>} بيضة باليوم،>}\\
&
\multicolumn{1}{r}{
\<ويبيع الدرزن بـ \LR{<currency>} \LR{<num2>}. كم يطلع دخله من البيض في الإسبوع؟>}\\
\bottomrule
\end{tabular}
\caption{Examples from DAR-MGSM localizations of MWPs for EGY, JOR, KSA, MOR, UAE, and MSA.}
\label{tab:egg-farm}
\end{table*}
More recent work studies reasoning as trajectories through residual-stream geometry and decomposes representations using sparse autoencoders and superposition analyses \citep{elhage2022superposition,bricken2023monosemanticity,templeton2024scaling,manson2025curved,li2025rema}. While prior work has primarily used these methods to study reasoning, factual recall, and circuit-level behavior, little attention has been paid to multilingual or culturally grounded evaluation settings. Here we bridge this gap.

\begin{table*}[t]
\centering
\small
\tabcolsep4pt
\begin{tabular}{lcccccccc}
\toprule
 & \textbf{ENG} & \textbf{Translated} & \textbf{MSA} & \textbf{EGY} & \textbf{JOR} & \textbf{KSA} & \textbf{MOR} & \textbf{UAE} \\
\midrule
\texttt{\textless name\textgreater} &
Lloyd &
\<لويد> &
\<موسى> &
\<متولي> &
\<وليد> &
\<عبد الرحيم> &
\<إبراهيم> &
\<مؤمن> \\
&
&
\textit{Lloyd} &
\textit{Musa} &
\textit{Metwally} &
\textit{Walid} &
\textit{Abdul Rahim} &
\textit{Ibrahim} &
\textit{Mumin} \\
\addlinespace

\texttt{\textless location\textgreater} &
Toulouse &
\<تولوز> &
\<تولوز> &
\<القاهرة> &
\<اربد> &
\<الرياض> &
\<مراكش> &
\<العين> \\
&
&
\textit{Toulouse} &
\textit{Toulouse} &
\textit{Cairo} &
\textit{Irbid} &
\textit{Riyadh} &
\textit{Marrakech} &
\textit{Al-Ain} \\
\addlinespace

\texttt{\textless num1\textgreater} &
252 &
252 &
\<٢٥٢> &
\<٢٥٢> &
\<٢٥٢> &
\<٢٥٢> &
252 &
\<٢٥٢> \\
% \addlinespace

\texttt{\textless num2\textgreater} &
5 &
5 &
\<٥> &
\<٥> &
\<٥> &
\<٥> &
5 &
\<٥> \\
% \addlinespace

\texttt{\textless currency\textgreater} &
dollar &
\<دولار> &
\<دولار> &
\<جنيه> &
\<دينار> &
\<ريال> &
\<درهم> &
\<درهم> \\
&
&
\textit{dollar} &
\textit{dollar} &
\textit{pound} &
\textit{dinar} &
\textit{riyal} &
\textit{dirham} &
\textit{dirham} \\
\bottomrule
\end{tabular}
\caption{Entity realizations across English, translated Arabic, MSA, EGY, JOR, KSA, MOR, and UAE.}
\label{tab:entities}
\end{table*}

\section{Dataset Curation} \label{sec:data_curation}

 We extended MGSM-Rev2 \cite{cobbe2021gsm8k, shi2022languagemodelsmultilingualchainofthought, peter2025mindgapnottranslation}, to include MSA\footnote{Note that the MSA translations of the entire GSM8K dataset by \citet{alenezi2026mathar} were done in parallel to our work and published in May 2026.} and five Arabic Dialects: Egyptian, Emirati, Jordanian, Moroccan, Saudi. MGSM-Rev2 comprises 250 selected MWPs from the GSM8K English benchmark \cite{cobbe2021gsm8k}. The English versions were the source text for translation for all the multilingual variants in MGSM and MGSM-Rev2 and also represent our starting point. The MWP from MGSM-Rev2 in Figure~\ref{fig:intro-example} demonstrates the US-centric nature of the original benchmark data. We therefore undertake both a \textit{translation} and a \textit{localization} approach to curating the data. Below, we describe both versions: standalone and separately.

\subsection{Manual Translation}
We hired professional translators to translate the MGSM-Rev2 250 MWPs into MSA and the five Arabic dialects.
Both a dedicated translator and a reviewer were assigned to each Arabic variety (see Appendix~\ref{sec:annotators_demog} for additional information on the annotators). As an additional quality check, native speakers conducted an in-house review of the professional translations. Correction rates ranged from 4.0\% for Saudi Arabic to 49.6\% for MSA, mostly due to misinterpretation of English expressions encoding mathematical relationships (See Appendix~\ref{sec:annotators_demog}).

The translation guidelines emphasized that entities such as places, names, and currencies should remain in the original form, i.e.,~should not be localized (see examples in Table~\ref{tab:entities}). 
 Additionally, we advised that dialectal translations should be natural with minimal overlap with MSA. For borrowed words, we asked annotators to adhere to linguistic forms used naturally in their dialect. For instance, the word for \textit{email} should be \setarab\<ايميل>. We had also allowed for spelling variation, where the meaning remained intact (see translation guidelines in Appendix~\ref{app:guidelines}).
The final translated dataset consists of 250 MWPs per dialect, i.e.,~a total of 1,500 MWPs covering six varieties.

\subsection{Localization}
Since direct translation would fail to incorporate cultural alignment with the dialects, we augmented the translated data with a semi-automated localization step, by replacing math question's entities with region-specific equivalents. We identified four entity types to be localized: names, locations, currencies, and numerals. See Table~\ref{tab:entities} for examples of entity replacements. 

The localizable entities were semi-automatically replaced throughout the MSA dataset with placeholders (e.g.,~\texttt{[person\_m]} for a human male name). Anchoring these placeholders to the translated text maintains structural and logical integrity, which is critical for such tasks. When multiple entities of the same category appeared in a single MWP, we assigned a unique index \(i\) to each entity variable to distinguish between different instances within the same category. 
As an example, if a MWP contained more than one male name, we assigned each name an index (e.g.,~\texttt{[person\_m\_1]}, \texttt{[person\_m\_2]}). This indexing scheme preserves the distinctions between the entities while maintaining consistent placeholder types. We then projected the placeholders onto the dialectal translations using a structured heuristic. 
\begin{table*}[t]
\small
\centering
\tabcolsep3pt
\renewcommand{\arraystretch}{1.2}
\begin{tabular}{c ll  ll  ll ll  ll ll l|lcc}
\toprule
 & \multicolumn{2}{c}{\textbf{MSA}} & \multicolumn{2}{c}{\textbf{EGY}} & \multicolumn{2}{c}{\textbf{JOR}} & \multicolumn{2}{c}{\textbf{KSA}} & \multicolumn{2}{c}{\textbf{MOR}} & \multicolumn{2}{c}{\textbf{UAE}} &&& \multicolumn{2}{c}{\textbf{Average}}  \\
 \midrule
 & \multicolumn{1}{c}{\textbf{T}} & \multicolumn{1}{c}{\textbf{L}} & \multicolumn{1}{c}{\textbf{T}} & \multicolumn{1}{c}{\textbf{L}} & \multicolumn{1}{c}{\textbf{T}} & \multicolumn{1}{c}{\textbf{L}} & \multicolumn{1}{c}{\textbf{T}} & \multicolumn{1}{c}{\textbf{L}} & \multicolumn{1}{c}{\textbf{T}} & \multicolumn{1}{c}{\textbf{L}} & \multicolumn{1}{c}{\textbf{T}} & \multicolumn{1}{c}{\textbf{L}} & &&\multicolumn{1}{c}{\textbf{T}} & \multicolumn{1}{c}{\textbf{L}} \\
 \toprule
Atlas-Chat-2B & \textbf{28.8} & 23.6 & \textbf{\underline{22.8}} & \underline{13.2} & \textbf{24.8} & 20.8 & \textbf{21.6} & 18.4 & \textbf{24.4} & 23.6 & \textbf{28.0} & 20.0 &&& \textbf{25.1} & 19.9 \\
Nile-Chat-4B & \textbf{\underline{38.0}} & \underline{28.4} & \textbf{35.2} & 32.4 & \textbf{\underline{36.8}} & \underline{31.2} & \textbf{\underline{35.2}} & \underline{24.4} & \textbf{15.2} & 14.8 & \textbf{36.4} & 36.0 &&& \textbf{32.8} & 27.9 \\
Atlas-Chat-9B & \textbf{66.0} & 61.6 & \textbf{55.2} & 52.0 & \textbf{61.6} & 58.4 & \textbf{56.4} & 55.2 & 58.4 & \textbf{60.0} & \textbf{62.4} & 61.2 &&& \textbf{60.0} & 58.1 \\
c4ai-r7b & \textbf{72.4} & 69.2 & \textbf{67.2} & 62.4 & \textbf{68.0} & 65.6 & \textbf{\underline{62.8}} & \underline{57.2} & 54.4 & \textbf{55.2} & 71.6 & \textbf{72.8} &&& \textbf{66.1} & 63.7 \\
Jais2-8B-Chat & 68.8 & 68.8 & \textbf{\underline{66.4}} & \underline{53.2} & \textbf{70.8} & 65.6 & \textbf{\underline{62.0}} & \underline{54.8} & \textbf{56.4} & 54.4 & \textbf{70.0} & 67.2 &&& \textbf{65.7} & 60.7 \\
Qwen2.5-7B & \textbf{75.6} & 69.6 & \textbf{69.6} & 63.6 & 72.0 & 72.0 & 62.0 & \textbf{65.2} & \textbf{38.8} & 38.4 & \textbf{76.8} & 74.0 &&& \textbf{65.8} & 63.8 \\
Gemma4-2B & \textbf{73.6} & 71.6 & 66.8 & \textbf{69.2} & \textbf{67.6} & 66.8 & \textbf{63.2} & 61.6 & \textbf{50.4} & 45.2 & 73.2 & \textbf{74.0} &&& \textbf{65.8} & 64.7 \\
Gemma4-4B & \textbf{84.4} & 81.2 & \textbf{80.0} & 79.6 & 82.8 & \textbf{84.0} & \textbf{75.2} & 73.2 & 66.4 & \textbf{66.8} & \textbf{85.6} & 83.2 &&& \textbf{79.1} & 78.0 \\
Gemma4-26B & 85.6 & \textbf{86.0} & \textbf{88.0} & 87.6 & \textbf{90.8} & 90.4 & 78.4 & \textbf{80.0} & \textbf{80.4} & 79.6 & \textbf{91.6} & 90.0 &&& \textbf{85.8} & 85.6 \\
Qwen-3.6 & \textbf{86.4} & 84.8 & 87.2 & \textbf{87.6} & \textbf{91.2} & 90.4 & \textbf{78.0} & 76.8 & 79.2 & \textbf{80.0} & \textbf{91.2} & 90.0 &&& \textbf{85.5} & 84.9 \\
GPT-5.4 & \textbf{90.0} & 89.2 & 89.2 & \textbf{90.0} & 93.2 & \textbf{94.8} & 85.6 & \textbf{86.4} & 84.4 & \textbf{84.8} & 92.0 & \textbf{93.2} &&& 89.1 & \textbf{89.7} \\
\hline
 &&  &  & & & & & & & & & &&& & \\
\textbf{Average} & \textbf{70.0} & 66.7 & \textbf{66.1} & 62.8 & \textbf{69.1} & 67.3 & \textbf{61.9} & 59.4 & \textbf{55.3} & 54.8 & \textbf{70.8} & 69.2 &&& \textbf{65.5} & 63.4 \\
\bottomrule
\end{tabular}
\caption{Accuracy (\%) of 11 models on translated (\textbf{T}) and localized (\textbf{L}) problems for MSA and five dialects. Bold: the higher of T and L. Underlined: T and L differ significantly (McNemar's test, $p<0.05$).}

\label{tab:all_results}
\end{table*}
The final phase of localization involved an automatic mapping of the variable placeholders to region-specific replacements. We replaced the names and the locations with regionally relevant counter-parts. We also adjusted the financial units to reflect country-specific currencies. Additionally, we localized numerical values and percentage signs to the Arabic-Indic numeral system for all dialects except Moroccan which uses Arabic numeral system. This ensured that both the linguistic and the mathematical components of the MWP aligned culturally with the target regions (see examples in Tables~\ref{tab:egg-farm} and~\ref{tab:entities}). A total of 51 MWPs did not have entities requiring localization (Appendix~\ref{sec:replacements} for more details). The dataset went through a final human validation step. The name category had the most issues with 47 wrong name replacement across all dialects (less than 1\% for all replacements). For example, our automatic scripts assumed that a name would appear once in the math problem, which led to failed replacements in cases where a name was mentioned twice or more.
We show the surface differences induced by each localization scheme in Table~\ref{tab:text-difference}.

\section{Experimental Setup}
\label{sec:exp_setup}

We experimented with several Arabic-centric and multilingual LLMs on their ability to answer MWPs from our dataset (for more details, see Appendix~\ref{app:models}). We evaluated the following multilingual models: Qwen-3.6, Gemma4 models, and GPT-5.4. We further evaluated the following Arabic-centric ones on dialectal models such as Atlas-Chat, NileChat, and Jais that provide support for different dialects. We used OpenRouter~\cite{openrouter2026} where applicable, otherwise relying on a local GPU for inference. The models were prompted as follows:\\
\begin{small}
\noindent\texttt{Solve the provided math problem. End with `The answer is: [number]'}.
\end{small}

We then extracted the numerical values and computed accuracy against a reference answer. Results were normalized by removing thousand separators. We note that Moroccan uses periods as thousand separators while other dialects use commas. 
We experimented with individual replacements. We manually audited model outputs to assess whether answer-extraction failures contributed to the reported errors. Section \ref{sec:err_anal} describes the audit, and Appendix~\ref{app:error_analysis} provides the category breakdown.

\begin{figure*}[t]
    \centering
    \includegraphics[trim={0cm 0cm 0cm 0cm}, clip, width=1.7\columnwidth]{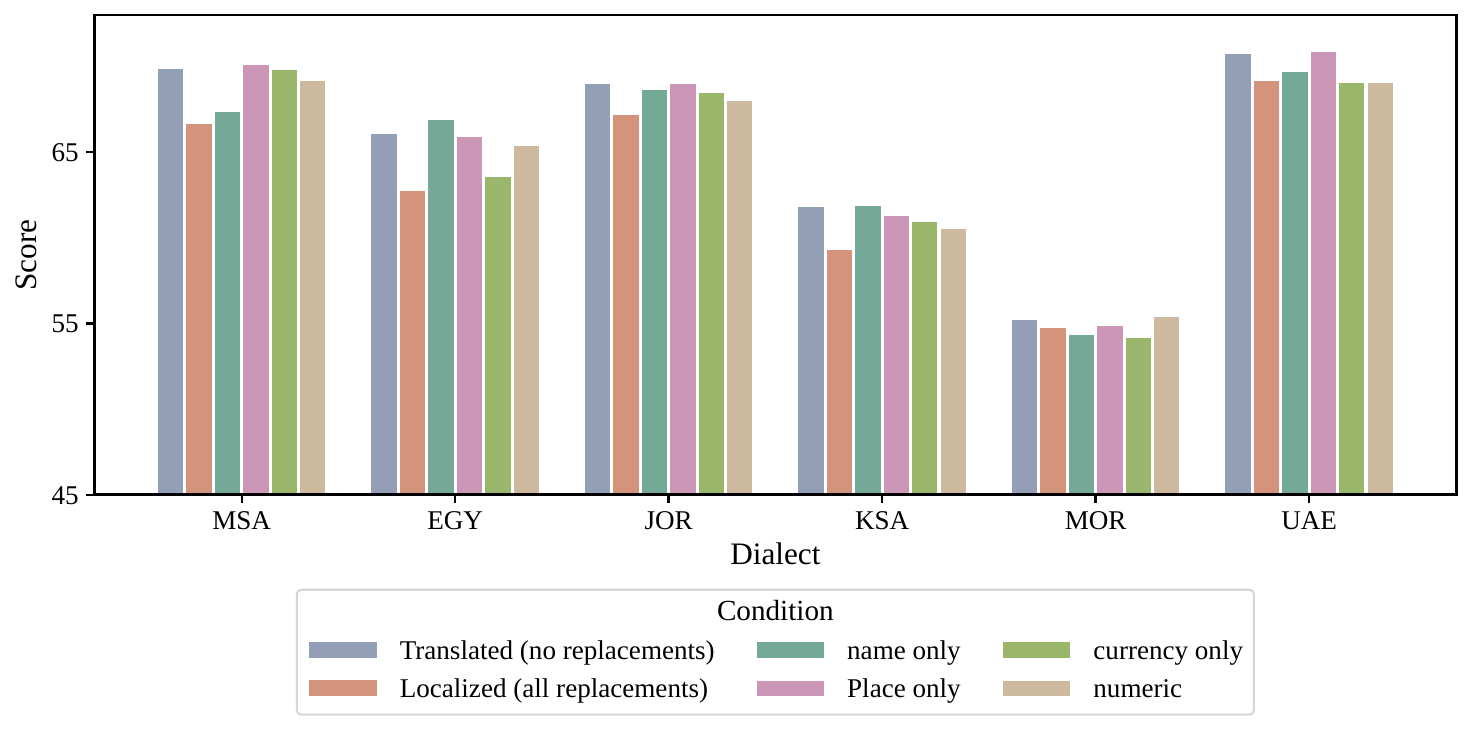}
    \caption{Performance by region and localization.}
    \label{fig:loc_chart}
\end{figure*}

\section{Results}\label{sec:results}
\subsection{Benchmark Results}
The overarching takeaway is that cultural localization exposes model fragility that direct translation conceals.
\paragraph{Overall evaluation results on DAR-MGSM} Table~\ref{tab:all_results} shows that the accuracy of Arabic-centric and multilingual models degrades when localized entities are introduced. This degradation can vary by model size and dialect. Individual replacements show only slight changes compared to all replacements. 

\paragraph{Degradation can vary depending on model size} While most models perform worse on localized data, flagship models like GPT~5.4 and Gemma~4 only show slight degradation. We observe that performance improves as the model size increases.

\paragraph{Localizations can impact a model's ability to correctly answer a MWP} In Figure~\ref{fig:loc_chart}, we also observe that across models, the accuracy drops for all Arabic varieties.  Performance appears to vary depending on the type of culturally-aware replacement (see Tables~\ref{tab:local_rep_results} and~\ref{tab:local_rep_results_specific}). For example, Mary vs Fatema, dirhams vs \$ or 'dollar', and so on. This reveals that adding localizations can impact on the model's ability to answer the MWP correctly.

\paragraph{Certain dialects are more vulnerable} Results in Table~\ref{tab:local_rep_results} and Table~\ref{tab:local_rep_results_specific} also reveal consistent model under-performance across both the translated and localized Moroccan MWPs in particular.

The largest drops in accuracy between translation-only and localization occur for MSA and Egyptian (-3.3 points each).

\paragraph{Statistical significance varies across different models} Table~\ref{tab:all_results} shows that, for flagship models that are large size, the results are not statistically significant.\footnote{Using \newcite{McNemar1947}'s significance test.} This does not indicate invariance, as upon further investigation, the models do not incorrectly answer the same questions across the translated and the localized MWPs. In smaller models such as Nile-Chat \citep{shang-etal-2025-nile} and Atlas-Chat \citep{shang-etal-2025-atlas}, however, the differences between translated and localized are statistically significant. This can indicate that these Arabic-centric models are more sensitive to regional differences. One possible explanation is that localizing MWPs can introduce challenges for these smaller models, and may result in performance degradation. 
Thus, the accuracy results do not provide a unified result: while some models show little change when introducing localizations, the MWPs that the models could correctly answer can differ. We further investigated this behavior focusing on Gemma~4 models through interpretability study, to understand the models behavior under localization.

\vspace{-2mm}
\subsection{Interpretability}
\vspace{-1.5mm}
\label{sec:inter_results}

We investigate the impact of localizations on a model's ability to answer MWP correctly. Our hypothesis is that the model's internal residual-stream trajectory changes, and so does the answer.

\paragraph{Sampling robustness of localization effects}
\begin{figure*}[t]
\centering
\includegraphics[trim={0cm 0cm 0cm 0cm}, clip,width=.75\textwidth]{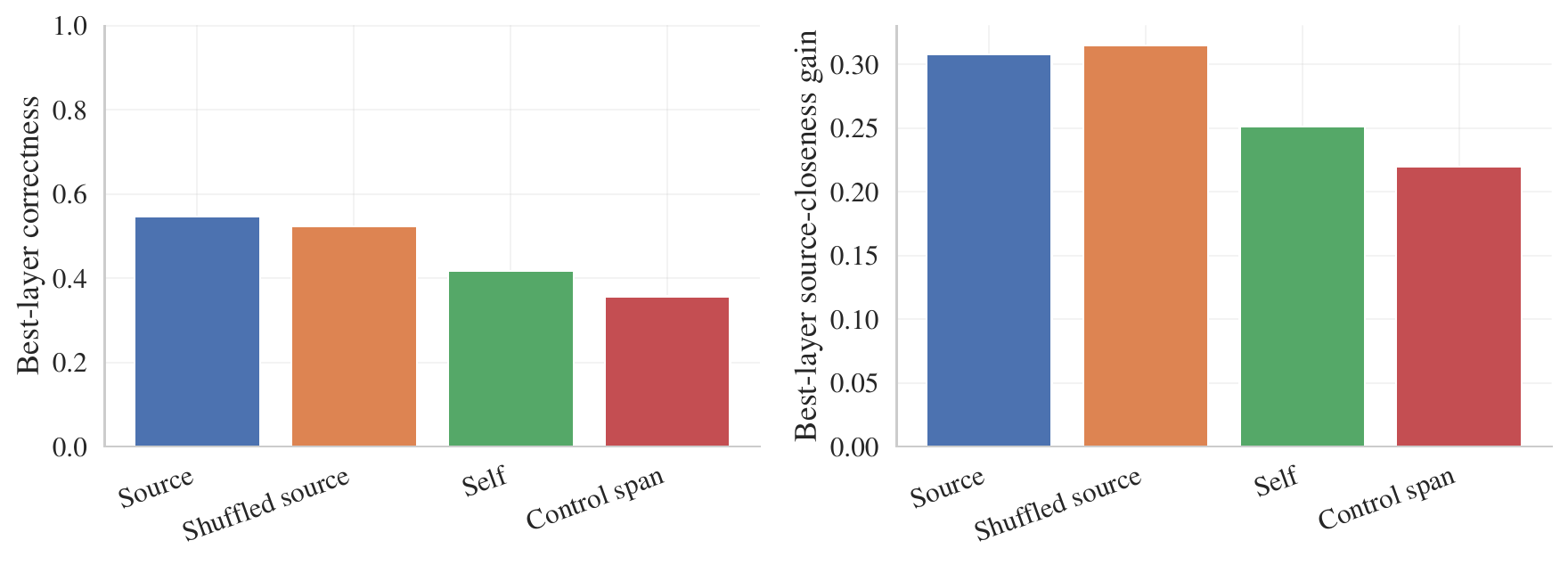}
\caption{Activation patching correctness by baseline. Source patching partially recovers source-condition behavior on discordant examples.}
\label{fig:patch-correctness}
\end{figure*}

 For each problem, we compare a baseline prompt (Translated) against localized variants (Localized) on the Gemma~4 models only for tractability. 
We divide examples into four groups: both correct, both wrong, baseline correct/localized wrong, and baseline wrong/localized correct. We refer to the latter two groups as \emph{discordant} examples. We compute Root Mean Square (RMS) trajectory distances by model, dialect, localization scheme, layer, and correctness group to whether localized and non-localized versions diverge when correctness changes.

We use activation patching to test whether localization-relevant activations causally drive the observed discordant behavioral shifts. We patch entity-span representations in both directions and check if the model’s answer and residual trajectory shift toward the source condition.
We compare source patching to controls: no patching, self-patching, shuffled non-localized sources, and patching an unrelated same-length span (See Appendix~\ref{app:activation-patching-method} and Figure~\ref{fig:patch-correctness}).
We then tested whether localization effects could be explained by ordinary stochastic decoding. 
We generated three samples for the translated condition and three for the localized condition and compared seed-to-seed residual-stream RMS distances 
against matched translated-to-localized RMS distances. 
\vspace{-1mm}
\paragraph{Localization changes which examples are solved}
Table~\ref{tab:all_results} shows that aggregate accuracy is broadly stable across localization schemes for Gemma~4 models. However, Table~\ref{tab:scheme-discordance} shows that paired outcomes are not stable. Under localization, models produce 316 discordant outcomes over 3,582 localizable pairs, corresponding to a discordance rate of 8.8\%, while the mean accuracy change is -1.2.
Thus, aggregate accuracy alone masks the effect: localization moves individual examples between the correct and the incorrect sets.

\paragraph{Smaller models are more sensitive, but larger models are not invariant} The discordance effect decreases with model scale. Table~\ref{tab:model-discordance} shows that, under full localization, Gemma4-2B has the highest discordance rate at $12.9\%$, while Gemma4-26B has the lowest at $5.6\%$. Larger models are therefore more stable under localization, but they exhibit paired examples where the baseline and localized prompts differ in correctness. This supports the idea that localizations can alter model predictions even for larger models. 
 \begin{table}[t]
  \centering
  \small

  \tabcolsep4pt
  \begin{tabular}{lcccc}

  \toprule
  \textbf{ } &
  \textbf{ } &
  \textbf{ } &
  \textbf{Discord.} &
  \textbf{Mean Acc.} \\
  \textbf{Model} &
  \textbf{$N$} &
  \textbf{Discord.} &
  \textbf{Rate} &
  \textbf{$\Delta$} \\
  \midrule
  Gemma4-2B  & 1,194 & 154 & 12.9\% & $-1.7$ pp \\
  Gemma4-4B  & 1,194 & 95  & 8.0\%  & $-1.8$ pp \\
  Gemma4-26B & 1,194 & 67  & 5.6\%  & $-0.3$ pp \\
  \bottomrule
  \end{tabular}

  \caption{Model-level paired discordance under full localization. Smaller models are more sensitive, 
  but
  localization effects remain present across scales. We exclude the 51 non-localizable MWPs that have differences in the numerals used and have slight discordance; however, we find it more interesting to study discordance among examples with bigger localization differences.}
  \label{tab:model-discordance}
  \end{table}
  
\paragraph{Correctness flips coincide with large residual-stream trajectory shifts}
We next ask whether paired behavioral changes correspond to changes in internal computation. Figure~\ref{fig:rms-correctness-group} and Table~\ref{tab:rms-correctness-group} summarize the main mechanistic result: examples that remain correct in both conditions have small residual-stream distances, while examples that flip correctness have much larger distances (Figure~\ref{fig:inteprepret_gemma4_main} support the same conclusion). Pairs correct in both baseline and localized forms have mean RMS around $0.29$. 

In contrast, pairs where only the baseline is correct have mean RMS around $1.50$, and pairs where only the localized version is correct have mean RMS around $1.50$. Pairs wrong in both conditions are intermediate, with mean RMS around $1.22$. This supports the hypothesis that when localization changes the answer, it also moves the model onto a different residual-stream trajectory.

Table~\ref{tab:sampling-robustness} shows that matched translated-to-localized residual-stream distances exceed within-translated seed-to-seed distances in 17 of 18 model--dialect cells (although Gemma4-2B MSA is very close to a tie at 1.004). The effect is most pronounced for Gemma4-26B, whose localization-to-sampling RMS ratios range from $1.07$ to $1.45$. Gemma4-2B shows smaller shifts, with ratios from $1.004$ to $1.214$, while Gemma4-4B is more variable and falls below $1$ only for MSA ($0.97$). Localization correctness-flip rates exceed sampling flip rates in 13 of 18 cells, are equal in one, and are lower in four. Thus, localization generally changes residual trajectories and correctness more than seed variation alone, although the magnitude depends on the model and dialect. These results suggest that the localization effect is not explained solely by stochastic decoding, but they do not independently establish a causal mechanism.
\paragraph{Activation patching supports a causal role for entity-span representations}
As shown in Figure~\ref{fig:patch-correctness}, RMS shifts show that answer flips coincide with trajectory changes, but do not establish causality. We therefore patch
  residual representations at the changed localization spans in both directions on discordant examples. Source patching
  outperforms the control-span baseline, reaching 54.7\% best-layer correctness and a 0.31
  best-layer source-closeness
  gain, compared with 35.7\% and 0.22 for unrelated-span patching. This supports a causal role for representations in shifting the target toward both the source answer and source residual trajectory, although it
  does not isolate an exclusively example-specific semantic mechanism.

\vspace{-1mm}
\paragraph{The effect is localized to entity spans, but not uniquely to matched entity identity}
The shuffled-source baseline is also strong, reaching $52.3\%$ best-layer correctness and a $0.31$ best-layer source-
  closeness gain, close to the matched source-patching results of $54.7\%$ and $0.31$. Because this control patches
  changed-span activations from mismatched examples, its performance suggests that recovery is not driven solely by
  transferring the exact matched name, location, currency, or numeral representation. We interpret the results
  as evidence that representations at localized spans are causally relevant, while the small gap between matched and
  shuffled sources leaves unresolved whether the operative signal encodes specific entity identity, a general source-condition representation, or a broader residual-state feature associated with those spans.

To summarize our results, these mechanistic shifts reveal that culturally adapting a MWP does not merely alter its surface vocabulary—it fundamentally changes the model's reasoning pathway. This proves that benchmarks relying solely on translated text may mask vulnerabilities in a model's real-world problem-solving capabilities.

\subsection{Error Analysis}\label{sec:err_anal}

To further understand the results, we perform manual analysis of errors in UAE and JOR by categorizing the errors into five categories: (a) Reasoning/setup error, when the model is providing incorrect interpretation from the beginning (e.g.,~units); (b) Final step error, where reasoning is correct but the final step is wrong.; (c) Missing step, when a model fails to include a necessary step.; (d) localization error, when the model struggles with localized entities; and (e) translation error. Among the analyzed errors, reasoning/setup errors such as misinterpreting the percentage/proportional relationships is the most frequent category, or incorrect interpretation of comparative expressions such as 'double'. 
We observed arithmetic errors, such as providing the wrong sum. In Figure~\ref{fig:err_JOR}, the first example 175 + 140 + 280 is 595 not 695.

Some examples produced alternative but internally consistent interpretations of the problem statement.
This could suggest that Some errors arise from ambiguity in the problem formulation rather than from incorrect reasoning. Interestingly, in the UAE dataset, we observe that incorrect MWP is because the reasoning language is Chinese (see examples in Appendix~\ref{app:error_analysis} from JOR and UAE).

\section{Conclusion and Future Work}

We introduced DAR-MGSM, the first manually curated benchmark for mathematical reasoning across five Arabic dialects and Modern Standard Arabic, with parallel translated-only and culturally localized versions. Our evaluation of eleven LLMs showed that localization generally reduces accuracy and can substantially change which problems models solve correctly, even when overall performance remains similar. Interpretability analysis linked these changes to shifts in internal representations. These findings highlight the importance of evaluating mathematical reasoning across both linguistic varieties and cultural contexts.

In future work, we plan to expand DAR-MGSM's coverage and to investigate methods for improving reasoning consistency across equivalent problem formulations. We also aim to extend this evaluation approach to other multilingual settings, supporting the development of educational AI that work reliably across languages and local contexts.

\section*{Limitations}
DAR-MGSM covers five Arabic dialects and Modern Standard Arabic, with results scoped to these varieties and the evaluated models. Cultural localization focuses on names, locations, currencies, and numeral conventions rather than all aspects of cultural context. Localization applies only to problems containing relevant entities, so comparisons for individual localization categories use the corresponding subsets.

\section*{Ethics}

DAR-MGSM is derived from MGSM-Rev2, which is distributed under the CC BY-SA 4.0 license. We use and adapt the source data in accordance with its license terms and provide attribution to its creators. The benchmark is intended to support research on mathematical reasoning across Arabic varieties and cultural contexts.

\bibliography{custom,camel-bib-v3}

@article{Ferguson:1959:diglossia,
	Author = {Charles A Ferguson},
	Journal = {Word},
	Number = {2},
	Pages = {325--340},
	Title = {{Diglossia}},
	Volume = {15},
	Year = {1959}}

@inproceedings{shi2022languagemodelsmultilingualchainofthought,
title     = {{Language Models Are Multilingual Chain-of-Thought Reasoners}},
  author    = {Shi, Freda and Suzgun, Mirac and Freitag, Markus and Wang, Xuezhi and Srivats, Suraj and Vosoughi, Soroush and Chung, Hyung Won and Tay, Yi and Ruder, Sebastian and Zhou, Denny and Das, Dipanjan and Wei, Jason},
  booktitle = {International Conference on Learning Representations},
  year      = {2023},
  url       = {https://mlanthology.org/iclr/2023/shi2023iclr-language/}
}

@inproceedings{alenezi2026mathar,
    title = "On {LLM} Prompting Techniques for {A}rabic Language Arithmetic Reasoning",
    author = "Alenezi, Reem  and
      Salman, Ayed Atallah",
    editor = "Al-Khalifa, Hend  and
      El-Haj, Mo  and
      Ezzini, Saad",
    booktitle = "The 7th Workshop on Open-Source {A}rabic Corpora and Processing Tools ({OSACT}7) with 5 Shared Tasks",
    month = may,
    year = "2026",
    address = "Palma, Mallorca (Spain)",
    publisher = "Association for Computational Linguistics",
    url = "https://aclanthology.org/2026.osact-1.13/",
    doi = "10.63317/52s6fyvjghcs",
    pages = "106--114"
}

@inproceedings{peter2025mindgapnottranslation,
    title = "Mind the Gap... or Not? {H}ow Translation Errors and Evaluation Details Skew Multilingual Results",
    author = "Peter, Jan-Thorsten  and
      Vilar, David  and
      Domhan, Tobias  and
      Malkin, Dan  and
      Freitag, Markus",
    editor = "Mille, Simon  and
      Gehrmann, Sebastian  and
      Schmidtov{\'a}, Patr{\'i}cia  and
      Du{\v{s}}ek, Ond{\v{r}}ej  and
      Fadaee, Marzieh  and
      Lo, Kyle  and
      Santus, Enrico  and
      Stanovsky, Gabriel",
    booktitle = "Proceedings of the Fifth Workshop on Generation, Evaluation and Metrics ({GEM})",
    month = jul,
    year = "2026",
    address = "San Diego, California, USA",
    publisher = "Association for Computational Linguistics",
    url = "https://aclanthology.org/2026.gem-main.22/",
    doi = "10.18653/v1/2026.gem-main.22",
    pages = "191--204",
    ISBN = "979-8-89176-423-1"
}

@inproceedings{huang2025benchmax,
    title = "{B}ench{MAX}: A {C}omprehensive {M}ultilingual {E}valuation {S}uite for {L}arge {L}anguage {M}odels",
    author = "Huang, Xu  and
      Zhu, Wenhao  and
      Hu, Hanxu  and
      He, Conghui  and
      Li, Lei  and
      Huang, Shujian  and
      Yuan, Fei",
    editor = "Christodoulopoulos, Christos  and
      Chakraborty, Tanmoy  and
      Rose, Carolyn  and
      Peng, Violet",
    booktitle = "Findings of the Association for Computational Linguistics: EMNLP 2025",
    month = nov,
    year = "2025",
    address = "Suzhou, China",
    publisher = "Association for Computational Linguistics",
    url = "https://aclanthology.org/2025.findings-emnlp.909/",
    doi = "10.18653/v1/2025.findings-emnlp.909",
    pages = "16751--16774",
    ISBN = "979-8-89176-335-7"
}

@InProceedings{alghamdi-liang-zhang:2022:LREC,
  author    = {Alghamdi, Reem  and  Liang, Zhenwen  and  Zhang, Xiangliang},
  title     = {{A}r{MATH}: a {D}ataset for {S}olving {A}rabic {M}ath {W}ord {P}roblems},
  booktitle      = {Proceedings of the Language Resources and Evaluation Conference},
  month          = {June},
  year           = {2022},
  address        = {Marseille, France},
  publisher      = {European Language Resources Association},
  pages     = {351--362},
  url       = {https://aclanthology.org/2022.lrec-1.37/}
}

@article{cobbe2021gsm8k,
  title={{T}raining {V}erifiers to {S}olve {M}ath {W}ord {P}roblems},
  author={Cobbe, Karl and Kosaraju, Vineet and Bavarian, Mohammad and Chen, Mark and Jun, Heewoo and Kaiser, Lukasz and Plappert, Matthias and Tworek, Jerry and Hilton, Jacob and Nakano, Reiichiro and Hesse, Christopher and Schulman, John},
  journal={arXiv preprint arXiv:2110.14168},
  year={2021},
  url={https://arxiv.org/abs/2110.14168}, 
}

@inproceedings{hoang-ha-2025-pensez,
    title = "Pensez: Moins de donn{\'e}es, meilleur raisonnement {--} Repenser les {LLM} fran{\c{c}}ais",
    author = "Hoang Ha, Huy",
    editor = "Bechet, Fr{\'e}d{\'e}ric  and
      Chifu, Adrian-Gabriel  and
      Pinel-sauvagnat, Karen  and
      Favre, Benoit  and
      Maes, Eliot  and
      Nurbakova, Diana",
    booktitle = "Actes des 32{\`e}me Conf{\'e}rence sur le Traitement Automatique des Langues Naturelles (TALN), volume 1 : articles scientifiques originaux",
    month = "6",
    year = "2025",
    address = "Marseille, France",
    publisher = "ATALA {\&} ARIA",
    url = "https://aclanthology.org/2025.jeptalnrecital-taln.35/",
    pages = "573--598",
    language = "fra"
}

@misc{openrouter2026,
  title        = {{O}pen{R}outer: Unified {API} for {L}arge {L}anguage {M}odels},
  author       = {{OpenRouter}},
  year         = {2026},
  howpublished = {\url{https://openrouter.ai}},
  note         = {Accessed: 2026-05-22}
}

@article{McNemar1947,
  doi = {10.1007/bf02295996},
  url = {https://doi.org/10.1007/bf02295996},
  year = {1947},
  month = jun,
  publisher = {Springer Science and Business Media {LLC}},
  volume = {12},
  number = {2},
  pages = {153--157},
  author = {Quinn McNemar},
  title = {{N}ote on the {S}ampling {E}rror of the {D}ifference {B}etween {C}orrelated {P}roportions or {P}ercentages},
  journal = {Psychometrika}
}

@techreport{elhage2021mathematical,
   title={{A Mathematical Framework for Transformer Circuits}},
   author={Elhage, Nelson and Nanda, Neel and Olsson, Catherine and Henighan, Tom and Joseph, Nicholas and Mann, Ben and Askell, Amanda and Bai, Yuntao and Chen, Anna and Conerly, Tom and DasSarma, Nova and Drain, Dawn and Ganguli, Deep and Hatfield-Dodds, Zac and Hernandez, Danny and Jones, Andy and Kernion, Jackson and Lovitt, Liane and Ndousse, Kamal and Amodei, Dario and Brown, Tom and Clark, Jack and Kaplan, Jared and McCandlish, Sam and Olah, Chris},
   year={2021},
   journal={Transformer Circuits Thread},
   institution = {Anthropic},
   url={https://transformer-circuits.pub/2021/framework/index.html}
}

@article{belrose2023tuned,
  author       = {Nora Belrose and
                  Zach Furman and
                  Logan Smith and
                  Danny Halawi and
                  Igor Ostrovsky and
                  Lev McKinney and
                  Stella Biderman and
                  Jacob Steinhardt},
  title        = {{E}liciting {L}atent {P}redictions from {T}ransformers with the {T}uned {L}ens},
  journal      = {CoRR},
  volume       = {abs/2303.08112},
  year         = {2023},
  url          = {https://doi.org/10.48550/arXiv.2303.08112},
  doi          = {10.48550/ARXIV.2303.08112},
  eprinttype   = {arXiv},
  eprint       = {2303.08112},
  bibsource    = {dblp computer science bibliography, https://dblp.org}
}

@inproceedings{meng2022locating,
author = {Meng, Kevin and Bau, David and Andonian, Alex and Belinkov, Yonatan},
 booktitle = {Advances in Neural Information Processing Systems},
 doi = {10.52202/068431-1262},
 editor = {S. Koyejo and S. Mohamed and A. Agarwal and D. Belgrave and K. Cho and A. Oh},
 pages = {17359--17372},
 publisher = {Curran Associates, Inc.},
 title = {Locating and Editing Factual Associations in {GPT}},
 url = {https://proceedings.neurips.cc/paper_files/paper/2022/file/6f1d43d5a82a37e89b0665b33bf3a182-Paper-Conference.pdf},
 volume = {35},
 year = {2022}
}

@techreport{olsson2022context,
   title={{In-context Learning and Induction Heads}},
   author={Olsson, Catherine and Elhage, Nelson and Nanda, Neel and Joseph, Nicholas and DasSarma, Nova and Henighan, Tom and Mann, Ben and Askell, Amanda and Bai, Yuntao and Chen, Anna and Conerly, Tom and Drain, Dawn and Ganguli, Deep and Hatfield-Dodds, Zac and Hernandez, Danny and Johnston, Scott and Jones, Andy and Kernion, Jackson and Lovitt, Liane and Ndousse, Kamal and Amodei, Dario and Brown, Tom and Clark, Jack and Kaplan, Jared and McCandlish, Sam and Olah, Chris},
   year={2022},
    institution = {Anthropic},
   journal={Transformer Circuits Thread},
   url={https://transformer-circuits.pub/2022/in-context-learning-and-induction-heads/index.html}
}

@inproceedings{nanda2023progress,
  title     = {{Progress Measures for Grokking via Mechanistic Interpretability}},
  author    = {Nanda, Neel and Chan, Lawrence and Lieberum, Tom and Smith, Jess and Steinhardt, Jacob},
  booktitle = {International Conference on Learning Representations},
  year      = {2023},
  url       = {https://mlanthology.org/iclr/2023/nanda2023iclr-progress/}
}

@techreport{elhage2022superposition,
   title={Toy Models of Superposition},
   author={Elhage, Nelson and Hume, Tristan and Olsson, Catherine and Schiefer, Nicholas and Henighan, Tom and Kravec, Shauna and Hatfield-Dodds, Zac and Lasenby, Robert and Drain, Dawn and Chen, Carol and Grosse, Roger and McCandlish, Sam and Kaplan, Jared and Amodei, Dario and Wattenberg, Martin and Olah, Christopher},
   year={2022},
   journal={Transformer Circuits Thread},
   institution={Anthropic},
   url={https://transformer-circuits.pub/2022/toy_model/index.html}
}

@techreport{bricken2023monosemanticity,
   title={Towards Monosemanticity: Decomposing Language Models With Dictionary Learning},
   author={Bricken, Trenton and Templeton, Adly and Batson, Joshua and Chen, Brian and Jermyn, Adam and Conerly, Tom and Turner, Nick and Anil, Cem and Denison, Carson and Askell, Amanda and Lasenby, Robert and Wu, Yifan and Kravec, Shauna and Schiefer, Nicholas and Maxwell, Tim and Joseph, Nicholas and Hatfield-Dodds, Zac and Tamkin, Alex and Nguyen, Karina and McLean, Brayden and Burke, Josiah E and Hume, Tristan and Carter, Shan and Henighan, Tom and Olah, Christopher},
   year={2023},
   journal={Transformer Circuits Thread},
   institution={Anthropic},
   url={https://transformer-circuits.pub/2023/monosemantic-features/index.html}
}

@techreport{templeton2024scaling,
   title={{S}caling {M}onosemanticity: {E}xtracting {I}nterpretable {F}eatures from {C}laude 3 {S}onnet},
   author={Templeton, Adly and Conerly, Tom and Marcus, Jonathan and Lindsey, Jack and Bricken, Trenton and Chen, Brian and Pearce, Adam and Citro, Craig and Ameisen, Emmanuel and Jones, Andy and Cunningham, Hoagy and Turner, Nicholas L and McDougall, Callum and MacDiarmid, Monte and Freeman, C. Daniel and Sumers, Theodore R. and Rees, Edward and Batson, Joshua and Jermyn, Adam and Carter, Shan and Olah, Chris and Henighan, Tom},
   year={2024},
   journal={Transformer Circuits Thread},
   institution={Anthropic},
   url={https://transformer-circuits.pub/2024/scaling-monosemanticity/index.html}
}

@misc{manson2025curved,
      title={{C}urved {I}nference: {C}oncern-{S}ensitive {G}eometry in {L}arge {L}anguage {M}odel {R}esidual {S}treams}, 
      author={Rob Manson},
      year={2025},
      eprint={2507.21107},
      archivePrefix={arXiv},
      primaryClass={cs.CL},
      url={https://arxiv.org/abs/2507.21107}, 
}

@misc{li2025rema,
      title={{REMA}: {A} {U}nified {R}easoning {M}anifold {F}ramework for {I}nterpreting {L}arge {L}anguage {M}odel}, 
      author={Bo Li and Guanzhi Deng and Ronghao Chen and Junrong Yue and Shuo Zhang and Qinghua Zhao and Linqi Song and Lijie Wen},
      year={2025},
      eprint={2509.22518},
      archivePrefix={arXiv},
      primaryClass={cs.AI},
      url={https://arxiv.org/abs/2509.22518}, 
}

@inproceedings{farsi-etal-2025-melac,
    title = "{MELAC}: {M}assive {E}valuation of {L}arge {L}anguage {M}odels with {A}lignment of {C}ulture in {P}ersian {L}anguage",
    author = "Farsi, Farhan  and
      Aghababaloo, Farnaz  and
      Motlagh, Shahriar Shariati  and
      Ghofrani, Parsa  and
      SadraeiJavaheri, MohammadAli  and
      Bali, Shayan  and
      Shabani, Amir Hossein  and
      Bijary, Farbod  and
      Zamaninejad, Ghazal  and
      Salehoof, AmirMohammad  and
      Momtazi, Saeedeh",
    editor = "Inui, Kentaro  and
      Sakti, Sakriani  and
      Wang, Haofen  and
      Wong, Derek F.  and
      Bhattacharyya, Pushpak  and
      Banerjee, Biplab  and
      Ekbal, Asif  and
      Chakraborty, Tanmoy  and
      Singh, Dhirendra Pratap",
    booktitle = "Proceedings of the 14th International Joint Conference on Natural Language Processing and the 4th Conference of the Asia-Pacific Chapter of the Association for Computational Linguistics",
    month = dec,
    year = "2025",
    address = "Mumbai, India",
    publisher = "The Asian Federation of Natural Language Processing and The Association for Computational Linguistics",
    url = "https://aclanthology.org/2025.ijcnlp-long.105/",
    doi = "10.18653/v1/2025.ijcnlp-long.105",
    pages = "1933--1950",
    ISBN = "979-8-89176-298-5"
}

@inproceedings{altakrori-etal-2026-dialectalarabicmmlu,
  title = {{D}ialectal{A}rabic{MMLU}: Benchmarking Dialectal Capabilities in {A}rabic and Multilingual Language Models},
  author = {Altakrori, Malik H. and Habash, Nizar and Lynn, Teresa and Samih, Younes and Freihat, Abed Alhakim and Chirkunov, Kirill and AbuOdeh, Muhammed and Florian, Radu and Nakov, Preslav and Aji, Alham Fikri},
  booktitle = {Proceedings of the Fifteenth Language Resources and Evaluation Conference (LREC 2026)},
  month = {May},
  year = {2026},
  pages = {3199--3219},
  address = {Palma, Mallorca, Spain},
  publisher = {European Language Resources Association (ELRA)},
  editor = {Piperidis, Stelios and Bel, Núria and van den Heuvel, Henk and Ide, Nancy and Krek, Simon and Toral, Antonio},
  doi = {10.63317/3cy68duew55b}
}

@inproceedings{saji2026reasoning,
    title = "{T}he {R}easoning {L}ingua {F}ranca: {A} {D}ouble-{E}dged {S}word for {M}ultilingual {AI}",
    author = "Saji, Alan  and
      Dabre, Raj  and
      Kunchukuttan, Anoop  and
      Puduppully, Ratish",
    editor = "Demberg, Vera  and
      Inui, Kentaro  and
      Marquez, Llu{\'i}s",
    booktitle = "Proceedings of the 19th Conference of the {E}uropean Chapter of the {A}ssociation for {C}omputational {L}inguistics (Volume 2: Short Papers)",
    month = mar,
    year = "2026",
    address = "Rabat, Morocco",
    publisher = "Association for Computational Linguistics",
    url = "https://aclanthology.org/2026.eacl-short.25/",
    doi = "10.18653/v1/2026.eacl-short.25",
    pages = "329--344",
    ISBN = "979-8-89176-381-4"
}

@inproceedings{sobhani2026mathmist,
    title = "{M}ath{M}ist: A Parallel Multilingual Benchmark Dataset for Mathematical Problem Solving and Reasoning",
    author = "E Sobhani, Mahbub  and
      Sayeedi, Md. Faiyaz Abdullah  and
      Mohiuddin, Tasnim  and
      Islam, Md Mofijul  and
      Shatabda, Swakkhar",
    editor = "Demberg, Vera  and
      Inui, Kentaro  and
      Marquez, Llu{\'i}s",
    booktitle = "Findings of the {A}ssociation for {C}omputational {L}inguistics: {EACL} 2026",
    month = mar,
    year = "2026",
    address = "Rabat, Morocco",
    publisher = "Association for Computational Linguistics",
    url = "https://aclanthology.org/2026.findings-eacl.131/",
    doi = "10.18653/v1/2026.findings-eacl.131",
    pages = "2524--2550",
    ISBN = "979-8-89176-386-9"
}

@inproceedings{chen-etal-2024-breaking,
    title = "{B}reaking {L}anguage {B}arriers in {M}ultilingual {M}athematical {R}easoning: {I}nsights and {O}bservations",
    author = "Chen, Nuo  and
      Zheng, Zinan  and
      Wu, Ning  and
      Gong, Ming  and
      Zhang, Dongmei  and
      Li, Jia",
    editor = "Al-Onaizan, Yaser  and
      Bansal, Mohit  and
      Chen, Yun-Nung",
    booktitle = "Findings of the Association for Computational Linguistics: EMNLP 2024",
    month = nov,
    year = "2024",
    address = "Miami, Florida, USA",
    publisher = "Association for Computational Linguistics",
    url = "https://aclanthology.org/2024.findings-emnlp.411/",
    doi = "10.18653/v1/2024.findings-emnlp.411",
    pages = "7001--7016"
}

@inproceedings{zhou2025culture,
    title = "{C}ulture is {N}ot {T}rivia: {S}ociocultural {T}heory for {C}ultural {NLP}",
    author = "Zhou, Naitian  and
      Bamman, David  and
      Bleaman, Isaac L.",
    editor = "Che, Wanxiang  and
      Nabende, Joyce  and
      Shutova, Ekaterina  and
      Pilehvar, Mohammad Taher",
    booktitle = "Proceedings of the 63rd Annual Meeting of the Association for Computational Linguistics (Volume 1: Long Papers)",
    month = jul,
    year = "2025",
    address = "Vienna, Austria",
    publisher = "Association for Computational Linguistics",
    url = "https://aclanthology.org/2025.acl-long.1256/",
    doi = "10.18653/v1/2025.acl-long.1256",
    pages = "25869--25886",
    ISBN = "979-8-89176-251-0"
}

@inproceedings{
rein2023gpqa,
title={{GPQA}: A {G}raduate-{L}evel {G}oogle-{P}roof {Q}\&{A} {B}enchmark},
author={David Rein and Betty Li Hou and Asa Cooper Stickland and Jackson Petty and Richard Yuanzhe Pang and Julien Dirani and Julian Michael and Samuel R. Bowman},
booktitle={Proceedings of the First Conference on Language Modeling: COLM 2024},
year={2024},
address={Philadelphia, PA, USA},
url={https://openreview.net/forum?id=Ti67584b98}
}

@inproceedings{hendrycks2021measuringmathematicalproblemsolving,
title     = {{M}easuring {M}athematical {P}roblem {S}olving with the {MATH} {D}ataset},
  author    = {Hendrycks, Dan and Burns, Collin and Kadavath, Saurav and Arora, Akul and Basart, Steven and Tang, Eric and Song, Dawn and Steinhardt, Jacob},
  year      = {2021},
  url       = {https://neurips.cc/virtual/2021/36901},  booktitle = {Proceedings of the Neural Information Processing Systems Track on Datasets and Benchmarks}
}

@inproceedings{hendrycks2021measuringmassivemultitasklanguage,
  title     = {{Measuring Massive Multitask Language Understanding}},
  author    = {Hendrycks, Dan and Burns, Collin and Basart, Steven and Zou, Andy and Mazeika, Mantas and Song, Dawn and Steinhardt, Jacob},
  booktitle = {International Conference on Learning Representations},
  year      = {2021},
  url       = {https://mlanthology.org/iclr/2021/hendrycks2021iclr-measuring/}
}

@article{amin2020teaching,
    author = {Tamer Amin and Diala Badreddine},
    title = {{Teaching science in Arabic: diglossia and discourse patterns in the elementary classroom}},
    journal = {International Journal of Science Education},
    volume = {42},
    number = {14},
    pages = {2290--2330},
    year = {2020},
    publisher = {Routledge},
    doi = {10.1080/09500693.2019.1629039},
    URL = {https://doi.org/10.1080/09500693.2019.1629039},
    eprint = {https://doi.org/10.1080/09500693.2019.1629039}
}

@incollection{amara1996arabic,
  title={{Arabic Diglossia in the Classroom Assumptions and Reality}},
  author={Amara, Muhammad Hasan},
  booktitle={Language and culture in the Near East},
  pages={131--142},
  year={1996},
  publisher={Brill},
  url={https://brill.com/edcollchap/book/9789004659377/B9789004648166_s009.xml}
}

@inproceedings{shang-etal-2025-nile,
    title = "{Nile-Chat: {E}gyptian Language Models for {A}rabic and {L}atin Scripts}",
    author = "Shang, Guokan  and
      Abdine, Hadi  and
      Chamma, Ahmad  and
      Mohamed, Amr  and
      Anwar, Mohamed  and
      Bounhar, Abdelaziz  and
      El Herraoui, Omar  and
      Nakov, Preslav  and
      Vazirgiannis, Michalis  and
      Xing, Eric P.",
    editor = "Darwish, Kareem  and
      Ali, Ahmed  and
      Abu Farha, Ibrahim  and
      Touileb, Samia  and
      Zitouni, Imed  and
      Abdelali, Ahmed  and
      Al-Ghamdi, Sharefah  and
      Alkhereyf, Sakhar  and
      Zaghouani, Wajdi  and
      Khalifa, Salam  and
      AlKhamissi, Badr  and
      Almatham, Rawan  and
      Hamed, Injy  and
      Alyafeai, Zaid  and
      Alowisheq, Areeb  and
      Inoue, Go  and
      Mrini, Khalil  and
      Alshammari, Waad",
    booktitle = "Proceedings of The Third Arabic Natural Language Processing Conference",
    month = nov,
    year = "2025",
    address = "Suzhou, China",
    publisher = "Association for Computational Linguistics",
    url = "https://aclanthology.org/2025.arabicnlp-main.25/",
    doi = "10.18653/v1/2025.arabicnlp-main.25",
    pages = "306--322",
    ISBN = "979-8-89176-352-4"
}

@inproceedings{shang-etal-2025-atlas,
    title = "{Atlas-Chat: Adapting Large Language Models for Low-Resource {M}oroccan {A}rabic Dialect}",
    author = "Shang, Guokan  and
      Abdine, Hadi  and
      Khoubrane, Yousef  and
      Mohamed, Amr  and
      Abbahaddou, Yassine  and
      Ennadir, Sofiane  and
      Momayiz, Imane  and
      Ren, Xuguang  and
      Moulines, Eric  and
      Nakov, Preslav  and
      Vazirgiannis, Michalis  and
      Xing, Eric P.",
    editor = "Hettiarachchi, Hansi  and
      Ranasinghe, Tharindu  and
      Rayson, Paul  and
      Mitkov, Ruslan  and
      Gaber, Mohamed  and
      Premasiri, Damith  and
      Tan, Fiona Anting  and
      Uyangodage, Lasitha",
    booktitle = "Proceedings of the First Workshop on Language Models for Low-Resource Languages",
    month = jan,
    year = "2025",
    address = "Abu Dhabi, United Arab Emirates",
    publisher = "Association for Computational Linguistics",
    url = "https://aclanthology.org/2025.loreslm-1.2/",
    pages = "9--30"
}

\appendix
\clearpage
\begin{onecolumn}
\section{Annotator Demographics}
\label{sec:annotators_demog}
\begin{table}[h]
\centering
\begin{tabular}{ccccccccc}
\toprule
\textbf{ID} & \textbf{Dataset} & \textbf{Native} & \textbf{Residence} & \textbf{Age} & \textbf{Gender} & \textbf{Degree} & \textbf{Task} & \textbf{Background} \\
\midrule
A1 & MSA & EGY & EGY & 30s & F & PhD  & T & L, LA \\
A2 & MSA & UAE & UAE & 40s & F & MA & R & L, CL \\
A3 & MSA & PAL & UAE & 30s & F & PhD & V & CL \\
\midrule
A4 & EGY & EGY & EGY & 30s & F & PhD  & T & L, LA \\
A5 & EGY & EGY & UAE & 40s & F & MA & R & L, CL \\
A6 & EGY & PAL & UAE & 30s & F & PhD & V & CL \\
\midrule
A7 & JOR & JOR & JOR & 30s & M & BA  & T & L, LA \\
A8 & JOR & PAL & UAE & 30s & F & PhD & R & CL \\
A9 & JOR & PAL & UAE & 30s & F & PhD & V & CL \\
\midrule
A10 & KSA & KSA & KSA & 20s & F & BA & T & CS \\
A11 & KSA & KSA & KSA & 30s & F & BA & R & LA \\
A12 & KSA & PAL & UAE & 30s & F & PhD & V & CL \\
\midrule
A13 & MOR & MOR & MOR & 20s & M & MA & T & CS \\
A14 & MOR & MOR & MOR & 30s & M & BA & R  & LA \\
A15 & MOR & PAL & UAE & 30s & F & PhD & V & CL \\
\midrule
A16 & UAE & UAE & US & 40s & F & PhD & T   & LA \\
A17 & UAE & UAE & UAE & 20s & M & MS & R & LA \\
A18 & UAE & PAL & UAE & 30s & F & PhD & V & CL \\
\bottomrule
\end{tabular}
\caption{Annotator demographics and roles for the translation task. There are three roles: Translate (T), Review (R), and Validate (V). The annotator background experience includes: Linguistics (L), Linguistic Annotator (LA), Computational Linguist (CL) and Computer Science (CS). All annotators are native speakers of Arabic, and they worked on their own native dialect or a dialect spoken in a country where they resided in. Annotators were paid \$15/hr.}
\end{table}

\begin{table}[h]
\centering
\small
\begin{tabular}{lrr}
\toprule
\textbf{Variety} & \textbf{Corrected / Total} & \textbf{Rate} \\
\midrule
MSA      & 124/250 & 49.6\% \\
Egyptian &  41/250 & 16.4\% \\
Jordanian & 77/250 & 30.8\% \\
Moroccan &  90/250 & 36.0\% \\
Emirati  &  70/250 & 28.0\% \\
Saudi    &  10/250 &  4.0\% \\
\bottomrule
\end{tabular}
\caption{Number and percentage of problems requiring major
edits or retranslation during internal quality assurance after handling the translations by our external vendor. Counts refer to problems, rather than individual edits, and describe corrections during dataset construction. These figures describe corrections during dataset curation, rather than residual error rates in the final benchmark. We conducted an additional internal quality assurance pipeline.}
\label{tab:translation-corrections}
\end{table}

\end{onecolumn}
\clearpage
\begin{onecolumn}
    


# B Translation Guidelines

**These are the guidelines as they were presented to our annotators.**

## Math Translation Guidelines

This task involves translating math problems in English into Arabic dialects and reviewing them for accuracy, fluency and naturalness. Each translation task is performed by two translators: one responsible for the translation, the other for reviewing it. See below for the translator and reviewer tasks for clear instructions (see example sheet in the appendix below).

You are provided with a translation request in an Excel sheet with four columns as detailed below:

| Column | Content | Description |
|---|---|---|
| A | English source text | This is the original source text in English |
| B | Dialect Translation | This is to add the translation in your dialect |
| C | Reject/Accept (choose one from list) | Choose whether you accept or reject the translation from the provided list |
| D | Justification (If Column C is "Reject") | If you rejected the translation, please provide an explanation of why it was rejected |

**Responsibilities and task description (for more details check the general instructions):**

**The translator's task**

1. **Preparation**
   - Open the excel file provided to you and ensure it is titled in your dialect.
   - Ensure it has all four columns with the English source (Col A), Dialect Translation (Col B), Reject/Accept (Col C), and Justification (Col D).
2. **Translation**
   - Read the full sentence to ensure you understand it before translating.
   - Translate the math problem from English (Column A) to your native dialect (Column B).
   - Use simple, concise, and direct language, maintaining the original meaning and clarity of the question.
   - After finishing the translation, make sure to check for the following:
     - Is the translation clear?
     - Does it sound natural in my dialect?

**The reviewer's task**

1. **Review the translation**
   - Review the translation in Column B.
   - Decide whether to accept or reject the translation based whether it meets the following requirement:
     - The translation should accurately reflect the meaning of the English text.
     - The names of places/proper nouns/currency should be kept as is.
     - The translation should sound natural in your dialect, with minimal overlap with MSA.
     - Borrowed words should be used naturally. For instance, is the word ايميل used and not بريد الكتروني as applicable per dialect.
     - The translation should follow the grammatical rules of your dialect.
2. **Feedback**
   - If rejecting, provide a justification in Column D.

As a quality assurance step, we will review accepted translations randomly to check the quality of the review process. A satisfactory translation would be following the checklist and general instructions in the guidelines.

**General instructions:**

- ☐ Confidence in translation: Only work on a sheet if you are confident in your translation capabilities for each specific question. Do not translate in cases where you doubt your understanding of the question. In such cases, simply move to the next line to avoid unnecessary effort. Please add a note of explanation in the comment column.
- ☐ Support tools: It is acceptable to use dictionaries or term bases to find the most suitable term for a word in a sentence. However, **automatic translation is strictly forbidden** as these translated texts will be part of a scientific study that cannot be influenced by translation technology.
- ☐ Make sure that the final translation reads like a question-answer conversation. For example:

  EN: Salem goes on a hiking trip for 3 days. The length of the trail is 24 kilometers. If he walks the same number of kilometers a day, how many kilometers will he walk every day?

  MSA: يقوم سالم برحلة مشي على الاقدام تستغرق ٣ ايام. ويفترض انه سيمشي اجمالا مسافة قدرها ٢٤ كيلومترا. اذا كان يمشي نفس عدد الكيلومترات يوميا فكم عدد الكيلومترات التي سيمشيها في اليوم؟

  PA: سالم رايح رحلة لثلاث ايام. لنفرض انو رح يمشي ٢٤ كيلومتر خلال كل الرحلة. اذا مشى نفس عدد الكيلومترات يوميا قديش بطلع ماشي بكل يوم؟

- ☐ Do not try to *invent terms* to make your translation “appear dialectical”. This holds the same for terms that do not have an equivalent term in Arabic (e.g. house flipping), please provide a brief description.
- ☐ For borrowed words in your variety, use the borrowed word as is to mimic the real dialect use. For instance, if the word "email" is commonly used in its transliterated form in your dialect (not translated as "بريد إلكتروني"), use the form "إيميل".
- ☐ While spelling variation is expected, it is advised to maintain consistent spelling for recurring dialectal terms within one dialect, For example, in the Egyptian dialect, the future tense is usually marked with **ه** at the beginning of the verb, like **هكلمك** ("I will call you"). Some speakers, however, prefer to use **ح** instead of **ه**, as in **حكلمك.** Whichever variant is used should be consistent for the dialect. See more details on acceptable variants in the orthographic guidelines part below.
- ☐ Maintain consistency with the number format in the source dataset. Use digits when digits are provided, and words when numbers are written out (e.g., dozens, fractions, or dual forms), as this reflects the dataset’s structure. For instance, if you have the number one written out in digits as ‘1’, then keep it as is (even if this does not follow the Arabic language rules).
- ☐ Any numeral system (Hindu-Arabic (١،٢،٣...) or Arabic (1,2,3...)) is acceptable in this dataset, use whatever is natural for your writing. This holds true for punctuation marks as well. See the table below for acceptable number/ punctuation marks variants.
- ☐ Entities such as proper nouns (e.g. names of places, people, organisations, and products) and units of measurement such as currencies, metres, miles, etc. should remain as is, please do not localize them for this task.

**Orthographic Guidelines**

Since DA has no official orthography standards, Arabs writing in DA tend to use a set of variant spellings that either capture the utterance phonology primarily or are inspired by Modern Standard Arabic orthographic decisions (etymological or historical spelling) (Habash et al., 2018). These are considered dialectical variants, where they do not impact readability and comprehension, but a variant is used consistently in a dialect. In cases where a variation affects readability, comprehension, or changes the meaning, the variant should be considered a spelling error. For example, the word “سال” when written with a hamza (سأل) means *asked*, whereas when written without a hamza (سال), it means *spelled* or *leaked*. Since the hamza alters the meaning, using the wrong form would be treated as a spelling error.

The following is a motivating example from Habash et al., (2018) showing 20+ spellings of the Egyptian Arabic word /mabi’ulha:š/ ‘he does not say it’ and their frequencies from a Google Search (September 29, 2017). Since all of these variants clearly indicate the intended meaning, they are all acceptable variants for the purpose of our task at hand.  Note that the variations include different decisions on how to represent the glottal stop whose MSA cognate is /q/ ق, shortened long vowels, attaching or detaching the negation proclitic mA/m, etc.

| Arabic Orthography | Arabic Transliteration | Frequency |
|---|---|---|
| مبيقولهاش | *mbyqwlhAš* | ≈ 26,000 |
| ما بيقولهاش | *mA byqwlhAš* | ≈ 13,000 |
| مابقلهاش، مبقولهاش، مبقلهاش، ما بقلهاش، مابيقولهاش | *mAbqlhAš, mbqwlhAš, mbqlhAš, mA bqlhAš, mAbyqwlhAš* | ≤ 10,000 |
| مابقولهاش، ما بقولهاش، مبيقلهاش، ما بيقلهاش | *mAbqwlhAš, mA bqwlhAš, mbyqlhAš, mA byqlhAš* | ≤ 1,000 |
| مبئلهاش، مابيئولهاش، ما بيئولهاش، مابيؤلهاش | *mbŷlhAš, mAbyŷwlhAš, mA byŷwlhAš, mAbyŵlhAš* | ≤ 100 |
| ما بيؤلهاش، مابئلهاش، مبيئولهاش، ما بيئلهاش، مابئولهاش، ما بئلهاش، ما بؤلهاش، مبئولهاش، مبيؤلهاش، مابؤلهاش، مبؤلهاش | *mA byŵlhAš, mAbŷlhAš, mbyŷwlhAš, mA byŷlhAš, mAbŷwlhAš, mA bŷlhAš, mA bŵlhAš, mbŷwlhAš, mbyŵlhAš, mAbŵlhAš, mbŵlhAš* | ≤ 10 |

We present below a list of common orthographic phenomena that we consider as dialectal variants. Some of these are the result of letter shape similarity and other phonological similarities. Please note, that not all "similar letters" are acceptable variants: if they are visually similar but phonologically not similar (don't sound similar), they are most likely errors.

| Variant Set Name | Variant Characters | Dialectal Variants Examples |
|---|---|---|
| Ta Marbuta /Ha | ه / ة | قارة أفريقيا الغنيه / قارة أفريقيا الغنية |
| Alif Maqsura/Ya | ي/ى | إزيك يا على! / إزيك يا علي!<br>سلملي علي الجيران! / سلم لي على الجيران! |
| Alif Hamzas | أ/إ/آ/ا | اهلأ ياجدعآن / أهلا يا جدعان |
| Hamza Forms | ؤ/ئ/ء/إ/أ | علماؤهم / علماءهم / علمائهم |
| Etymological Variants | ق / أ/إ/ؤ/ئ/ء | أنا قلبي دليلي / أنا ألبي دليلي |
| Etymological Variants | ت / ط | هادا الفسطان حلو / هادا الفستان حلو |
| Etymological Variants | ت/ث | حبيتك كثير / حبيتك كتير |
| Etymological Variants | ز/ض | الزابط واقف هناك / الضابط واقف هناك |
| Etymological Variants | د/ذ/ز | دنب الكلب / ذنب الكلب<br>ذنبك على جنبك / زمبك على جمبك |
| Splits | و،ب،ك، ل، ح | ح نشوف و ح نعرف / حنشوف وحنعرف |
| Merges | ما ، يا، لا | ماشفت / ما شفت<br>يامحمد / يا محمد<br>لاتحكي / لا تحكي |
| Shortening & Lengthening | و/ ، ا/ ،ي/ | شوفت اللي عملاه ؟ شفت اللي عاملاه؟ |

| | | |
|---|---|---|
| | (long vowel spelling variant of short vowel spelling via elided diacritics) | |

It is not possible to create a full, exact list of acceptable variants for all possible contexts, as some contexts can lead the reader to a different unintended meaning. For example, while Ta Marbuta and Ta Mamduda may be interchangeable safely in contexts like جمعية الحارة / جمعيت الحاره, they may not be in limited contexts like حركة الجدعان / حركت الجدعان.

**Acceptable number/punctuation marks variants**

| | |
|---|---|
| 1,2,3,4...<br>%,<br>,<br>.<br>;<br>? | ١، ٢، ٣، ٤....<br>٪<br>،<br>,<br>؛<br>؟ |

**Appendix A:**

Screenshot of Excel sheet

| 1 | English source text | Dialect translation | Accept/Reject | Justification (If Column C is "Reject") |
|---|---|---|---|---|
| 2 | James decides to run 3 sprints 3 times a week. He runs 60 meters each sprint. How many total meters does he run a week? | جيمس قرر يركض ثلاث لفات ثلاث مرات بالاسبوع. كل لفة بركض فيها ستين متر. قديش عدد الامتار الي بركضها اسبوعيا؟ | Accept | |

\end{onecolumn}

\clearpage
\section{Localizations}
\label{sec:replacements}

\begin{table}[h]
    \centering
    \small
    \begin{tabular}{l|l|llllll}
        \toprule
         & \textbf{EN} & \textbf{MSA} & \textbf{EGY} & \textbf{JOR} & \textbf{KSA} & \textbf{MOR} & \textbf{UAE} \\
        \midrule
        Avg. local./MWP & 5.79 & 5.30 & 5.32 & 5.64 & 5.55 &  5.59 & 5.40\\
        \bottomrule
    \end{tabular}
    \caption{Number of localizations made per line on EN data compared to the other dialects.}
    \label{tab:placeholder}
\end{table}

\begin{table}[h]
  \centering
  \begin{tabular}{lrrrrr}
  \toprule
  Scheme & $N$ & Char diff & Word diff & Char edit & Word edit \\
  \midrule
  All      & 1194 & 10.8\% & 18.0\% & 19.1 & 6.2 \\
  Names    & 1190 & 6.3\%  & 7.2\%  & 10.9 & 2.4 \\
  Currency & 522  & 5.7\%  & 11.4\% & 10.6 & 4.2 \\
  Location & 173  & 2.2\%  & 1.5\%  & 4.4  & 0.5 \\
  Numerals & 1329 & 3.4\%  & 9.3\%  & 5.9  & 3.1 \\
  \bottomrule
  \end{tabular}
  \caption{Average surface difference by localization scheme. Localization edits are small and targeted rather than
broad rewrites.}
  \label{tab:text-difference}
\end{table}

\section{Additional Details on the Experimental Setup}
\label{app:models}

\begin{table*}[!h]
\small
\centering
\setlength{\tabcolsep}{3.5pt}
\begin{tabular}{@{}rllccll@{}}
\toprule
 & \textbf{Model} & \textbf{Identifier} & \textbf{Ar} & \textbf{En} & \textbf{Access} & \textbf{Licence} \\
\midrule
1  & Atlas-Chat-2B & \texttt{MBZUAI-Paris/Atlas-Chat-2B}               & $\bullet$ &           & Local      & Gemma ToU \\
2  & Nile-Chat-4B  & \texttt{MBZUAI-Paris/Nile-Chat-4B}                & $\bullet$ &           & Local      & Gemma ToU \\
3  & Atlas-Chat-9B & \texttt{MBZUAI-Paris/Atlas-Chat-9B}               & $\bullet$ &           & Local      & Gemma ToU \\
4  & c4ai-r7b      & \texttt{CohereLabs/c4ai-command-r7b-arabic-02-2025} & $\bullet$ & $\bullet$ & Local      & CC BY-NC 4.0 \\
5  & Jais2-8B-Chat & \texttt{inception42/Jais-2-8B-Chat}               & $\bullet$ & $\bullet$ & Local      & Apache 2.0 \\
6  & Qwen2.5-7B    & \texttt{Qwen/Qwen2.5-7B-Instruct}                 & $\bullet$ & $\bullet$ & Local      & Apache 2.0 \\
7  & Gemma4-2B     & \texttt{google/gemma-4-E2B-it}                    & $\bullet$ & $\bullet$ & Local      & Apache 2.0 \\
8  & Gemma4-4B     & \texttt{google/gemma-4-E4B-it}                    & $\bullet$ & $\bullet$ & Local      & Apache 2.0 \\
9  & Gemma4-26B    & \texttt{google/gemma-4-26B-A4B-it}                & $\bullet$ & $\bullet$ & Local      & Apache 2.0 \\
10 & Qwen-3.6      & \texttt{qwen/qwen3.6-plus}                        & $\bullet$ & $\bullet$ & OpenRouter & API terms \\
11 & GPT-5.4       & \texttt{openai/gpt-5.4}                           & $\bullet$ & $\bullet$ & OpenRouter & API terms \\
\bottomrule
\end{tabular}
\caption{Evaluated models, in the order of Table~\ref{tab:all_results}. Identifiers refer to the Hugging Face Hub (local models) or OpenRouter (API models). Ar and En mark the languages each model supports (Arabic, English). Gemma ToU: Google's Gemma Terms of Use, inherited from the Gemma base models of Atlas-Chat and Nile-Chat. Gemma4-2B and Gemma4-4B are the E2B and E4B variants (2B and 4B effective parameters); Gemma4-26B is a mixture-of-experts model with 4B active parameters.}
\label{tab:List_models}
\end{table*}
\begin{twocolumn}
    
\paragraph{Decoding}
Reasoning (thinking) is disabled for GPT-5.4, Qwen-3.6 and the Gemma~4 models. All local models and Qwen-3.6 use greedy decoding (temperature 0). OpenRouter does not pass a temperature setting to GPT-5.4, so GPT-5.4 uses the provider's default sampling. Each model is run once per variety and condition; no scores are averaged over repeated runs. Responses that reach the token limit are scored as generated.

\paragraph{Prompting}
Except for the Gemma~4 models, every model receives the instruction in Section~\ref{sec:exp_setup} as the system message and the problem as the user message, formatted with the model's own chat template; when a template has no system role, the instruction is prepended to the user message. These models generate at most 1,024 new tokens.

\paragraph{Gemma~4}
The Gemma~4 models use the pipeline of the interpretability analysis (Section~\ref{sec:inter_results}) with greedy decoding: at most 512 new tokens, the system prompt ``You are a helpful assistant.'', and an instruction to end with ``Final answer: [number]''.

\paragraph{Hardware}
The Arabic-centric models and Qwen2.5-7B run in Hugging Face Transformers in bf16 without quantization on a single NVIDIA A100 40GB GPU. The Gemma~4 models run on a machine with 4 Intel Xeon Gold 6150 CPU cores (2.70~GHz) and a single NVIDIA Tesla V100-SXM2 32GB GPU. GPT-5.4 and Qwen-3.6 are queried through OpenRouter, at a total cost of approximately \$100.

\paragraph{Run-to-run variation}
In MSA, the place-only and currency-only questions are identical to the translated ones, so any
difference between these three conditions reflects run-to-run variation, not localization.
Seven of the eleven models give identical answers in all three. For the other four, accuracy
varies by at most 0.8 points (2.8 for Qwen-3.6), and the average over all models varies by only
0.2 points.
\end{twocolumn}

\twocolumn[{%
\section{Interpretability}
\label{sec:interpret_main}
\centering
\includegraphics[width=\textwidth]{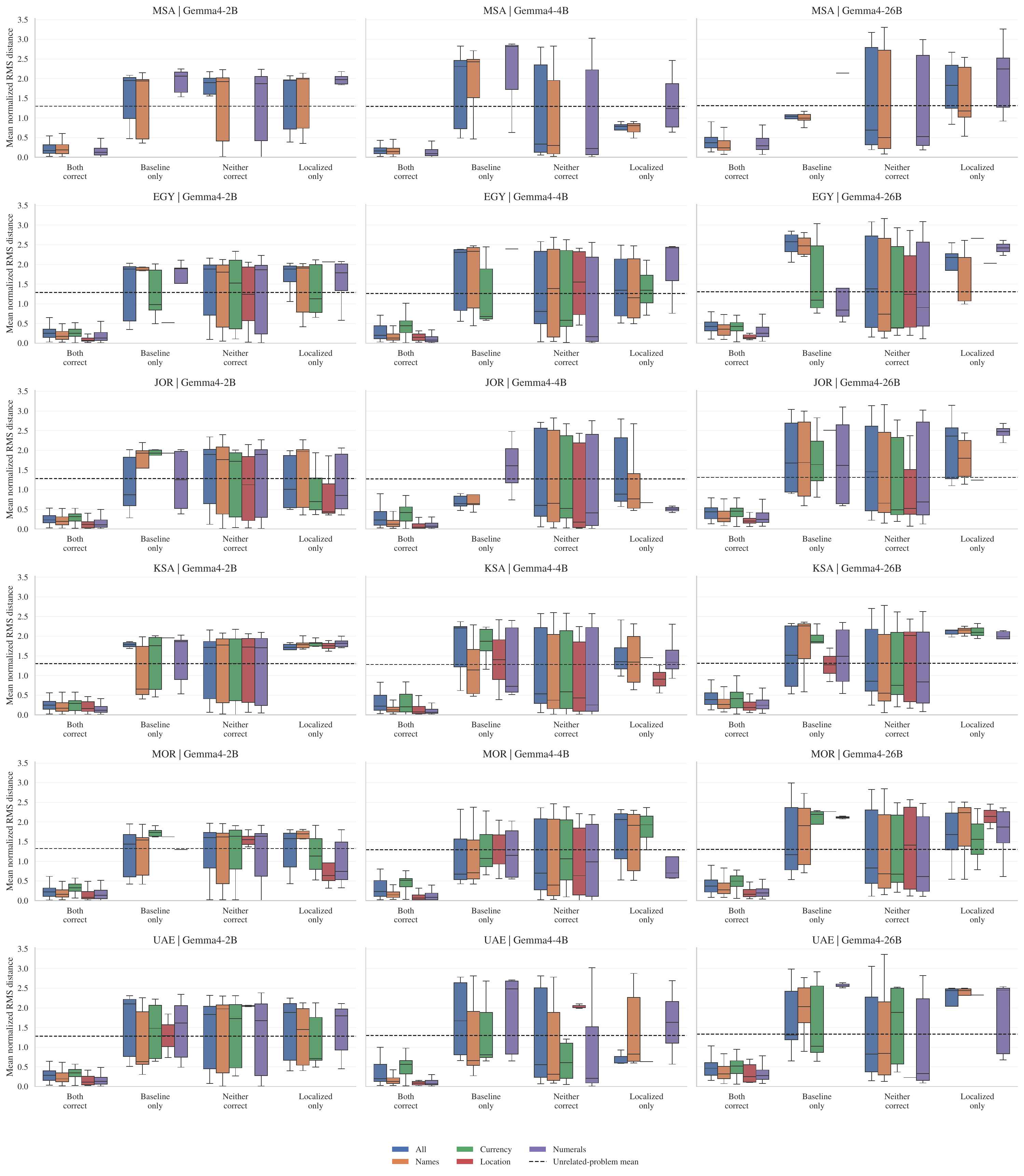}
\captionof{figure}{Residual-stream path distance by correctness group; dashed line = unrelated-problem average.}
\label{fig:inteprepret_gemma4_main}
\vspace{1.5em}
}]
 
\clearpage
\begin{onecolumn}
\section{Results}
\label{sec:loc_chart}

\begin{table}[h]
\centering
\begin{tabular}{lcccccc}
\toprule
 & \textbf{T} & \multicolumn{1}{l}{\textbf{L}} & \multicolumn{1}{l}{\textbf{Name}} & \multicolumn{1}{l}{\textbf{Curr.}} & \multicolumn{1}{l}{\textbf{Loc.}} & \multicolumn{1}{l}{\textbf{Num.}} \\
 \toprule
\textbf{MSA} & 70.0 & 66.7 & 67.5 & 69.9 & 70.1 & 69.2 \\
\textbf{EGY} & 66.1 & 62.8 & 66.9 & 63.6 & 66.0 & 65.5 \\
\textbf{JOR} & 69.1 & 67.3 & 68.7 & 68.5 & 69.1 & 68.1 \\
\textbf{KSA} & 61.9 & 59.4 & 62.0 & 61.0 & 61.4 & 60.6 \\
\textbf{MOR} & 55.3 & 54.8 & 54.4 & 54.2 & 54.9 & 55.5 \\
\textbf{UAE} & 70.8 & 69.2 & 69.7 & 69.1 & 70.9 & 69.1\\
\bottomrule
\end{tabular}
\caption{Model evaluation average results across MSA and the dialects: EGY, JOR, KSA, MOR and UAE. We report accuracy for the different localizations schemes over the 250 problems. \textbf{T} denotes Translated data; \textbf{L} denotes Translated and fully Localized data, \textbf{Name} denotes  name replacements only, \textbf{Curr.} denotes currency replacements only, \textbf{Loc.} denotes location replacements only and \textbf{Num.} denotes numerical value replacements only.} %\textcolor{red}{done}} 
\label{tab:local_rep_results}
\end{table}

\begin{table}[h]
\centering
\begin{tabular}{lcccc}
\toprule
 & \multicolumn{1}{l}{\textbf{Name}} & \multicolumn{1}{l}{\textbf{Curr.}} & \multicolumn{1}{l}{\textbf{Loc.}} & \multicolumn{1}{l}{\textbf{Num.}} \\
 \toprule
\textbf{MSA} & 65.7 & 71.0 & 75.8 & 69.2 \\
\textbf{EGY} & 66.7 & 67.2 & 78.8 & 65.5 \\
\textbf{JOR} & 67.0 & 64.8 & 78.8 & 68.1 \\
\textbf{KSA} & 61.9 & 60.6 & 58.6 & 60.6 \\
\textbf{MOR} & 53.7 & 56.3 & 55.6 & 55.5 \\
\textbf{UAE} & 68.6 & 70.3 & 84.8 & 69.1\\
\bottomrule
\end{tabular}
\caption{Model evaluation average results across MSA and the dialects: EGY, JOR, KSA, MOR and UAE. We report accuracy for the different localizations schemes. Each scheme is evaluated only on the questions it applies to (localizable entities): \textbf{Name}, \textbf{Curr.}, \textbf{Loc.} and \textbf{Num.}.} 
\label{tab:local_rep_results_specific}
\end{table}

\begin{table}[h]
\centering
\begin{tabular}{lrr}
\toprule
\textbf{Correctness group} & \textbf{$N$} & \textbf{Mean RMS} \\
\midrule
Baseline only correct   & 556  & 1.50 \\
Localized only correct  & 470  & 1.50 \\
Neither correct         & 2722 & 1.22 \\
Both correct            & 9476 & 0.29 \\
\bottomrule
\end{tabular}
\caption{Mean residual-stream RMS distance by paired correctness group,
pooled across the three Gemma4 models, six dialects, and localization
schemes. Discordant examples exhibit substantially larger trajectory shifts
than examples that are correct in both conditions.}
\label{tab:rms-correctness-group}
\end{table}

\begin{table}[!h]
\centering
\begin{tabular}{lrrrrr}
\toprule
\textbf{Scheme} & \textbf{$N$} & \textbf{Discordant}
& \textbf{Discord.} & \textbf{mean}
& \textbf{Localized only} / \textbf{} \\
\textbf{} & \textbf{} & \textbf{}
& \textbf{rate} & \textbf{acc. $\Delta$}
& \textbf{} \textbf{Baseline only} \\
\midrule
All      & 3582  & 316 & 8.8\% & $-1.2$ pp & 136 / 180 \\
Names    & 3570  & 311 & 8.7\% & $-0.2$ pp & 152 / 159 \\
Currency & 1566  & 153 & 9.8\% & $-3.0$ pp & 53 / 100 \\
Location & 519   & 29  & 5.6\% & $+0.2$ pp & 15 / 14 \\
Numerals & 3987  & 217 & 5.4\% & $+0.3$ pp & 114 / 103 \\
\midrule
Total    & 13224 & 1026 & 7.8\% & $-0.6$ pp & 470 / 556 \\
\bottomrule
\end{tabular}

\caption{Paired discordance by localization scheme, pooled across the three
Gemma4 models and six dialects. Only eligible, quality-filtered pairs are
included for each scheme, so $N$ varies by scheme. The final column reports
cases where only the localized prompt is correct and where only the
translated baseline is correct; these totals match the discordant-group
sizes in Table~\ref{tab:rms-correctness-group}. Accuracy $\Delta$ is
localized minus translated accuracy in percentage points.}
\label{tab:scheme-discordance}
\end{table}

\begin{table}[h!]
\centering
\begin{tabular}{llrrr}
\toprule
\textbf{Model} & \textbf{Dialect} & \textbf{Loc./S.\ RMS}
& \textbf{Samp.\ flip} & \textbf{Loc.\ flip} \\
\midrule
            & MSA & 1.004 & 16.0\% & 12.0\% \\
            & EGY & 1.025 & 12.0\% & 13.3\% \\
Gemma4-2B   & \textbf{JOR} & \textbf{1.214} & 4.0\% & 14.7\% \\
            & KSA & 1.024 & 9.3\% & 11.3\% \\
            & MOR & 1.111 & 13.3\% & 18.7\% \\
            & UAE & 1.043 & 9.3\% & 12.0\% \\
\midrule
            & \emph{MSA} & \emph{0.965} & 4.0\% & 6.7\% \\
            & EGY & 1.140 & 4.0\% & 8.0\% \\
Gemma4-4B   & JOR & 1.091 & 4.0\% & 6.0\% \\
            & KSA & 1.083 & 16.0\% & 12.0\% \\
            & MOR & 1.158 & 8.0\% & 8.0\% \\
            & UAE & 1.153 & 5.3\% & 4.7\% \\
\midrule
            & MSA & 1.197 & 2.7\% & 3.3\% \\
            & EGY & 1.259 & 5.3\% & 6.0\% \\
Gemma4-26B  & JOR & 1.073 & 6.7\% & 4.0\% \\
            & KSA & 1.221 & 9.3\% & 11.3\% \\
            & MOR & 1.182 & 4.0\% & 4.7\% \\
            & \textbf{UAE} & \textbf{1.449} & 1.3\% & 5.3\% \\
\bottomrule
\end{tabular}

\caption{Sampling robustness under full localization. Each model--dialect
cell uses up to 50 matched problems and three generation seeds.
Loc./S.\ RMS is the ratio of matched translated-to-localized RMS distance to
within-translated seed-to-seed RMS distance, computed from unrounded mean
values. Values above $1$ indicate that localization changes residual
trajectories more than sampling variation in the translated condition: this
holds in 17 of 18 cells (smallest margin 1.004 for Gemma4-2B/MSA); the only
exception is Gemma4-4B/MSA (\emph{0.965}, italicized). \textbf{Bold} marks
the largest ratio within each model. Flip rates report correctness changes
across the corresponding paired comparisons; localization flip rates exceed
sampling flip rates in 13 of 18 cells, equal them in 1, and fall below in 4.}
\label{tab:sampling-robustness}
\end{table}
\end{onecolumn}

\clearpage
\begin{figure*}[h]
\centering
\includegraphics[trim={0cm 0cm 0cm 0cm}, clip, width=.80\textwidth, height=.60\textheight]{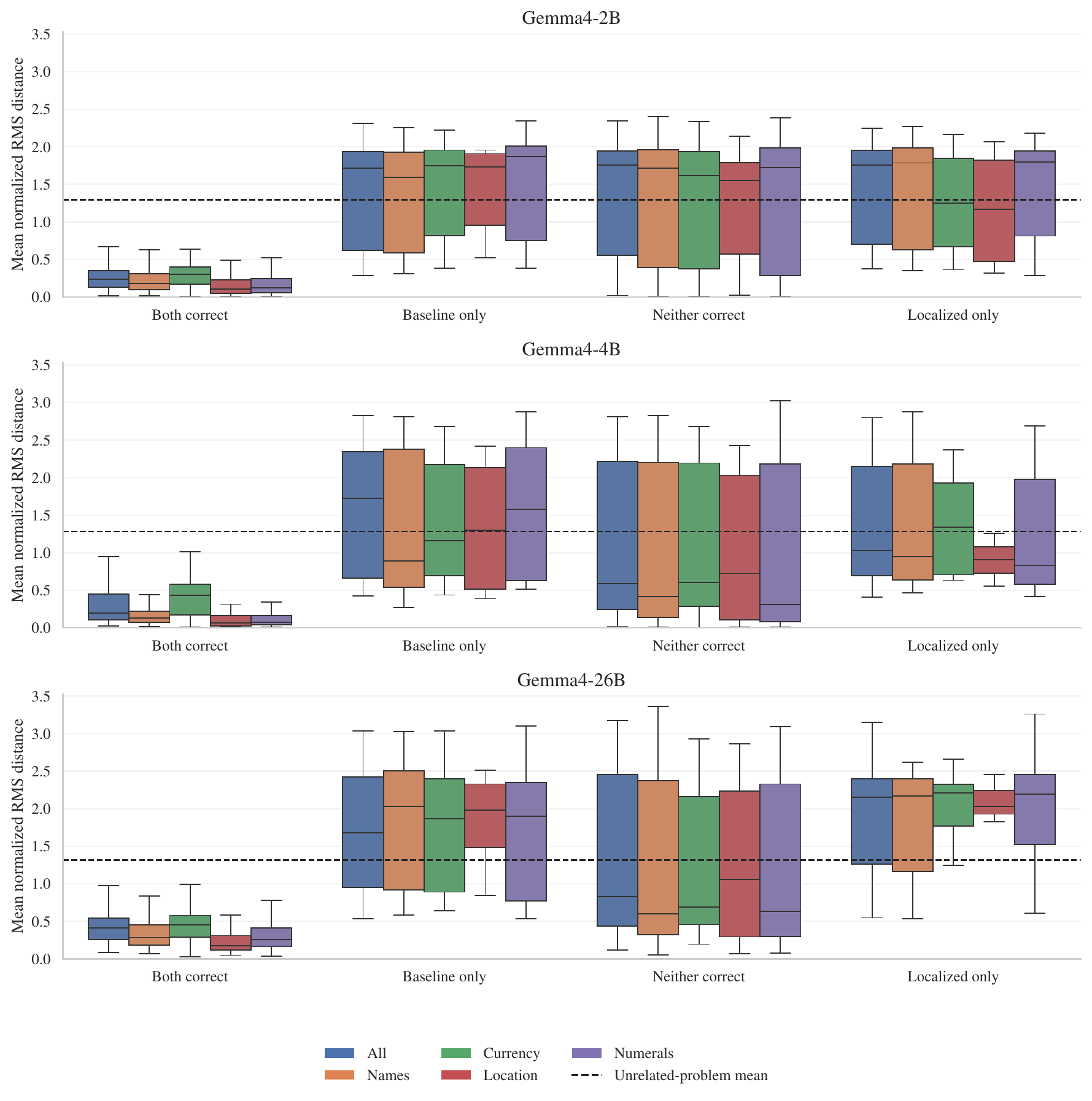}
\caption{Residual-stream RMS distance by paired correctness group. Discordant examples show substantially larger trajectory separation than examples that remain correct in both conditions.}
\label{fig:rms-correctness-group}
\end{figure*}

\clearpage
\twocolumn 

\section{Activation Patching Procedure}
  \label{app:activation-patching-method}
  We use activation patching to test whether correctness changes on discordant examples are causally linked to the localized entity tokens. A
  discordant example is one where the model answers correctly in one condition but incorrectly in the other, e.g., the translated non-localized
  prompt is correct while the localized prompt is wrong, or vice versa. For each such pair, we run patching in both directions: from the non-
  localized prompt into the localized prompt, and from the localized prompt into the non-localized prompt. If the localized entity
  representation is causally involved in the answer flip, then replacing the target prompt's entity-span residual stream with the source
  prompt's corresponding entity-span residual stream should move the patched generation toward the source condition, both in answer correctness
  and in residual-stream geometry.

  Let $x^{src}$ denote the source prompt and $x^{tgt}$ denote the target prompt. The source is the condition whose internal representation we
  patch from, and the target is the condition whose forward pass is modified. For each transformer layer $\ell$, we run the target prompt while
  replacing selected residual-stream vectors at selected token positions with the corresponding source residual-stream vectors:
  \[
  h^{tgt}_{\ell, p_i} \leftarrow h^{src}_{\ell, q_i},
  \]
  where $p_i$ is a target token position and $q_i$ is the matched source token position. We then decode from the patched target run and evaluate
  whether the patched answer becomes correct, whether it matches the source answer, and whether the patched residual trajectory moves closer to
  the source trajectory.

  \paragraph{Span identification.}
  For each localized/non-localized prompt pair, we first identify the changed text spans between the two problem statements. Rather than
  assuming a fixed name list, we compare the two strings and extract the non-matching contiguous spans in order of appearance. These spans are
  then paired by order: the first changed span in the localized prompt is paired with the first changed span in the non-localized prompt, the
  second with the second, and so on. This handles examples with multiple changed names, locations, or currencies without needing to decide
  manually which entity should correspond to which one. For source patching, we select the relevant localized entity span and its aligned non-
  localized span.

  \paragraph{Tokenization and multi-token spans.}
  After identifying character-level spans, we map each span to token positions using the model tokenizer and the final chat-formatted prompt. A
  single entity may correspond to one token or multiple tokens. For example, a name may be split into several subword tokens, and a location may
  occupy multiple Arabic or Latin-script tokens. We therefore patch spans as token sequences rather than as single tokens. If the source span
  tokenizes to positions
  \[
  (q_1, q_2, \ldots, q_m)
  \]
  and the target span tokenizes to positions
  \[
  (p_1, p_2, \ldots, p_n),
  \]
  we patch the aligned positions in order. When $m=n$, this is a direct one-to-one token replacement. When the token lengths differ, we patch
  the overlapping prefix up to $\min(m,n)$ tokens. Thus, the operation remains localized to the aligned entity span while avoiding invalid
  assumptions about tokenizer granularity. The recorded \texttt{source\_token\_count}, \texttt{target\_token\_count}, and
  \texttt{patch\_token\_count} allow us to audit these cases.

  \paragraph{Layer sweep.}
  We repeat the patch for each residual-stream layer from early to late layers. This gives a layer-wise causal profile: if a name or location
  representation matters only after it has been integrated into the model's reasoning state, patching should have stronger effects at those
  layers. For each layer, we generate an answer from the patched target run and compute both behavioral and geometric metrics.

  \paragraph{Behavioral metric.}
  The primary behavioral metric is patched correctness:
  \[
  \mathrm{Correct}^{patch}_{\ell} =
  \mathbb{1}\{\mathrm{answer}(x^{tgt}_{patched,\ell}) = y\},
  \]
  where $y$ is the gold answer. We also compare the patched answer to the clean source and clean target answers to see whether the patch moves
  the model toward the source condition.

  \paragraph{Geometric metric.}
  We also measure whether the patched residual trajectory moves closer to the source trajectory. Let $R^{src}$ be the residual-stream trajectory
  of the clean source run, $R^{tgt}$ the clean target trajectory, and $R^{patch}_{\ell}$ the patched target trajectory after patching at layer $
  \ell$. We compute RMS distances between trajectories:
  \[
  d(R^a, R^b)
  =
  \sqrt{
  \frac{1}{|\mathcal{L}|}
  \sum_{k \in \mathcal{L}}
  \left\|R^a_k - R^b_k\right\|_2^2
  }.
  \]
  The clean target-to-source distance $d(R^{tgt}, R^{src})$ is the unpatched baseline. The patched-to-source distance $d(R^{patch}_{\ell},
  R^{src})$ tells us whether the patch moved the target trajectory toward the source. We report a source-distance ratio:
  \[
  \mathrm{SourceRatio}_{\ell}
  =
  \frac{
  d(R^{patch}_{\ell}, R^{src})
  }{
  d(R^{tgt}, R^{src})
  },
  \]
  and a source-closeness gain:
  \[
  \mathrm{Gain}_{\ell}
  =
  1 - \mathrm{SourceRatio}_{\ell}.
  \]
  A positive gain means the patched run is closer to the source trajectory than the clean target was; a gain near zero means the patch did not
  move the trajectory; a negative gain means the patch moved it farther away.

  \paragraph{Patching methods and controls.}
  We compare the entity-span source patch against several controls:
  \begin{itemize}
      \item \texttt{source\_patch}: patch the matched entity span from the source condition into the target condition. This is the main causal
      intervention. If entity localization drives the correctness flip, this patch should move the target answer and residual trajectory toward
      the source condition.
      \item \texttt{no\_patch}: run the target condition without modifying activations. This gives the unpatched target behavior and residual
      trajectory.
      \item \texttt{self\_patch}: patch the target span using activations from the same target condition. This controls for the mechanics of
      patching itself; because the patched vectors come from the same condition, it should not systematically change behavior.
      \item \texttt{shuffled\_source}: patch from a source example that is not the matched problem. This tests whether any source-condition
      activation is sufficient, or whether the effect depends on the correct problem-matched entity representation.
      \item \texttt{control\_span}: patch an unrelated same-length span rather than the localized entity span. This tests whether the effect is
      specific to the entity tokens rather than caused by overwriting arbitrary positions in the prompt.
  \end{itemize}

  \paragraph{Interpretation.}
  The causal prediction is that \texttt{source\_patch} should outperform the controls on discordant examples: it should increase the probability
  of recovering the source-side correct answer and produce a positive source-closeness gain. Strong evidence requires that this effect is larger
  than \texttt{self\_patch}, \texttt{control\_span}, and \texttt{shuffled\_source}; otherwise the result could be explained by patching
  artifacts, arbitrary position overwrites, or generic source-condition activations rather than the localized entity representation itself.

\clearpage
\begin{onecolumn}

\section{Error Analysis Examples}
\label{app:error_analysis}
    
\vspace{-1.7cm}
\begin{figure*}[h!]
    \centering
    \makebox[\textwidth][c]{
        \includegraphics[trim={0.5cm 0.5cm 0.5cm 0.5cm}, clip, width=1.8\textwidth]{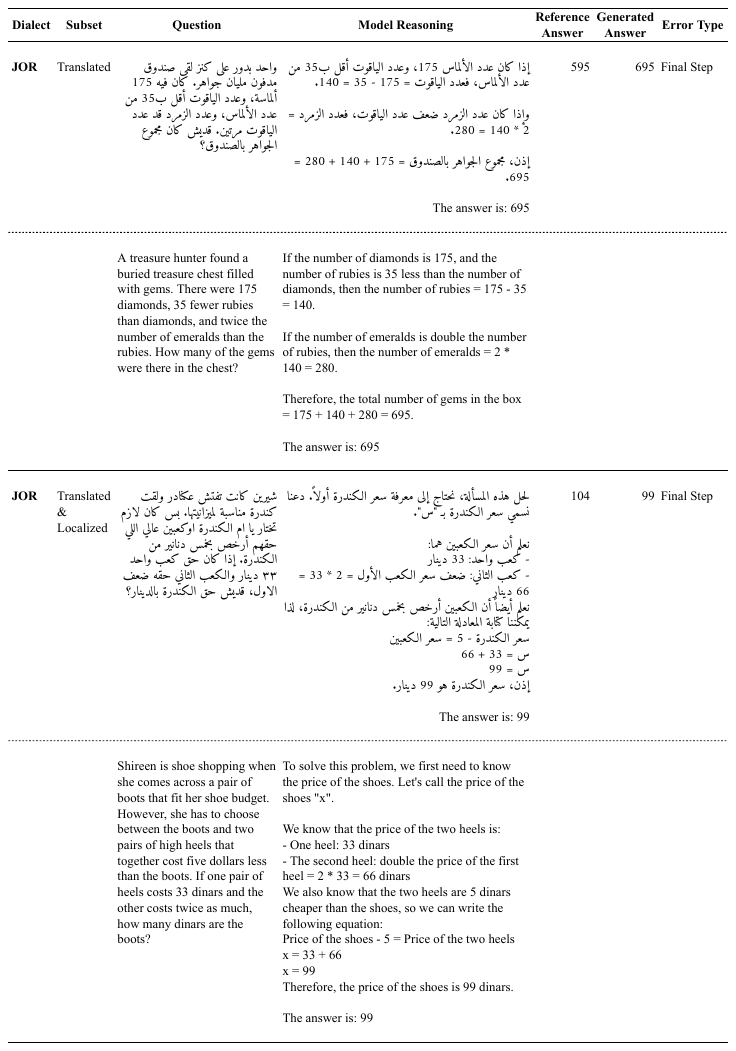}
    }
    \caption{Examples of errors in Jordanian Arabic.}
    \label{fig:err_JOR}
\end{figure*}

\clearpage
\begin{figure*}[h!]
    \centering
    \makebox[\textwidth][c]{
        \includegraphics[trim={0.5cm 0.5cm 0.5cm 0.5cm}, clip, width=1.8\textwidth]{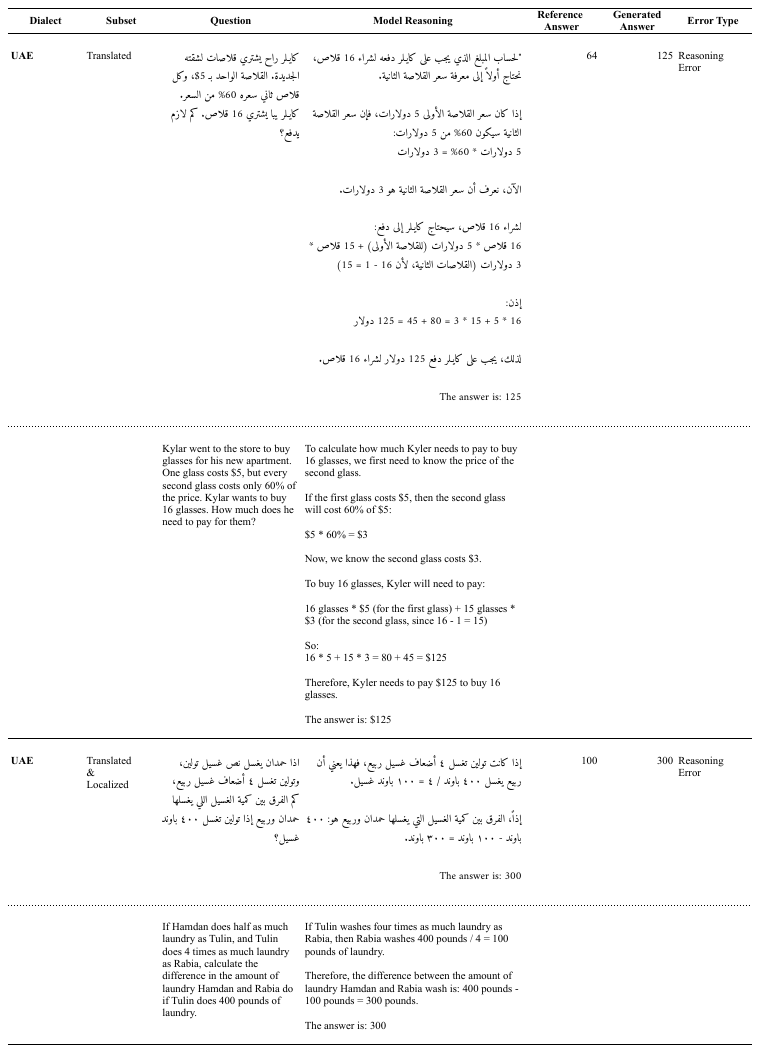}
    }
    \caption{Examples of errors in Emirati Arabic.}
    \label{fig:err_EMA}
\end{figure*}
\end{onecolumn}

\end{document}